\documentclass[10pt,journal,compsoc]{IEEEtran}
\usepackage[nocompress]{cite}

\usepackage[margin=0pt,font=small,labelfont=bf,labelsep=endash,tableposition=top]{caption}
\usepackage{booktabs}
\usepackage{bm}

\usepackage{amsmath}
\usepackage{array}
\usepackage{dsfont}
\usepackage{algorithm}  
\usepackage{algorithmic}

\usepackage[caption=false,font=footnotesize,labelfont=sf,textfont=sf]{subfig}
\usepackage{gensymb}
\usepackage{url}
\usepackage{graphicx}
\usepackage{amsmath}
\usepackage{amssymb}

\usepackage{epsfig}
\usepackage{epstopdf}
\usepackage{makecell,multirow,diagbox}
\usepackage{color}
\usepackage{soul}
\usepackage{url}
\usepackage{amsmath}
\usepackage{array}

\usepackage[utf8]{inputenc}
\usepackage{xcolor}
\usepackage[hidelinks]{hyperref}

\definecolor{yl_color}{RGB}{128, 255, 0}

\usepackage{mathtools,xspace}

\def\etal{{\it et al.}\xspace}
\hypersetup{
    colorlinks=true,
    breaklinks=false,
    urlcolor= red,
    linkcolor= red,
    bookmarksopen=false,
    filecolor=black,
    citecolor=blue,
    linkbordercolor=blue
}

\begin{document}

 \title{LRANet++: Low-Rank Approximation Network for Accurate and Efficient Text Spotting}

\author{Yuchen Su,
        Zhineng Chen,~\IEEEmembership{Member,~IEEE},
        Yongkun Du,
        Zuxuan Wu,~\IEEEmembership{Member,~IEEE},
        Hongtao Xie,
        Yu-Gang Jiang,~\IEEEmembership{Fellow,~IEEE}
\IEEEcompsocitemizethanks{
\IEEEcompsocthanksitem{This work was supported by National Natural Science Foundation of China under Grants 62427819 and 62172103. (Corresponding author: Zhineng~Chen) }
\IEEEcompsocthanksitem{Yuchen Su and Yongkun Du are with the College of Computer Science and Artificial Intelligence, Fudan University, Shanghai 200433, China (e-mail: ycsu23@m.fudan.edu.cn,  ykdu23@m.fudan.edu.cn).}
\IEEEcompsocthanksitem{Zhineng Chen, Zuxuan Wu, and Yu-Gang Jiang are with the Institute of Trustworthy Embodied AI, College of Intelligent Robotics and Advanced Manufacturing, Fudan University, Shanghai 200433, China (e-mail: zhinchen@fudan.edu.cn, zxwu@fudan.edu.cn, ygj@fudan.edu.cn).}
\IEEEcompsocthanksitem{Hongtao Xie is with the School of Information Science and Technology, University of Science and Technology of China, Hefei 230022, China (e-mail: htxie@ustc.edu.cn).}
}
}

\IEEEtitleabstractindextext{%

\begin{abstract}

End-to-end text spotting aims to jointly optimize text detection and recognition within a unified framework. Despite significant progress, designing an accurate and efficient end-to-end text spotter for arbitrary-shaped text remains challenging. We identify the primary bottleneck as the lack of a reliable and efficient text detection method. To address this, we propose a novel parameterized text shape representation based on low-rank approximation for precise detection and a triple assignment detection head for fast inference. Specifically, unlike current data-irrelevant shape representation methods, we exploit shape correlations among labeled text boundaries to construct a robust low-rank subspace. By minimizing an $\ell_1$-norm objective, we extract orthogonal vectors that capture the intrinsic text shape from noisy annotations, enabling precise reconstruction via the linear combination of only a few basis vectors. Next, the triple assignment scheme decouples training complexity from inference speed. It utilizes a deep sparse branch to guide an ultra-lightweight inference branch, while a dense branch provides rich parallel supervision. Building upon these advancements, we integrate the enhanced detection module with a lightweight recognition branch to form an end-to-end text spotting framework, termed LRANet++, capable of accurately and efficiently spotting arbitrary-shaped text. Extensive experiments on challenging benchmarks demonstrate the superiority of LRANet++ compared to state-of-the-art methods. Code is available at:  \url{https://github.com/ychensu/LRANet-PP}.
\end{abstract}

\begin{IEEEkeywords}
Scene text spotting, Low-rank approximation, Triple assignment
\end{IEEEkeywords}}

\maketitle

\IEEEdisplaynontitleabstractindextext
\IEEEpeerreviewmaketitle

\IEEEraisesectionheading{\section{Introduction}
\label{sec:introduction}}
\IEEEPARstart{D}{e}tecting and recognizing scene text simultaneously, \textit{a.k.a.} text spotting, has potential applications in various fields such as visual question answering, document image understanding, and multimodal retrieval. Despite significant progress, existing methods \cite{liu2021abcnet,huang2022swintextspotter,ye2023deepsolo,huang2023estextspotter} fail to achieve an ideal trade-off between accuracy and efficiency, limiting their applicability in many real-world scenarios.

\begin{figure}[t]
    \centering
    \includegraphics[width=\linewidth]{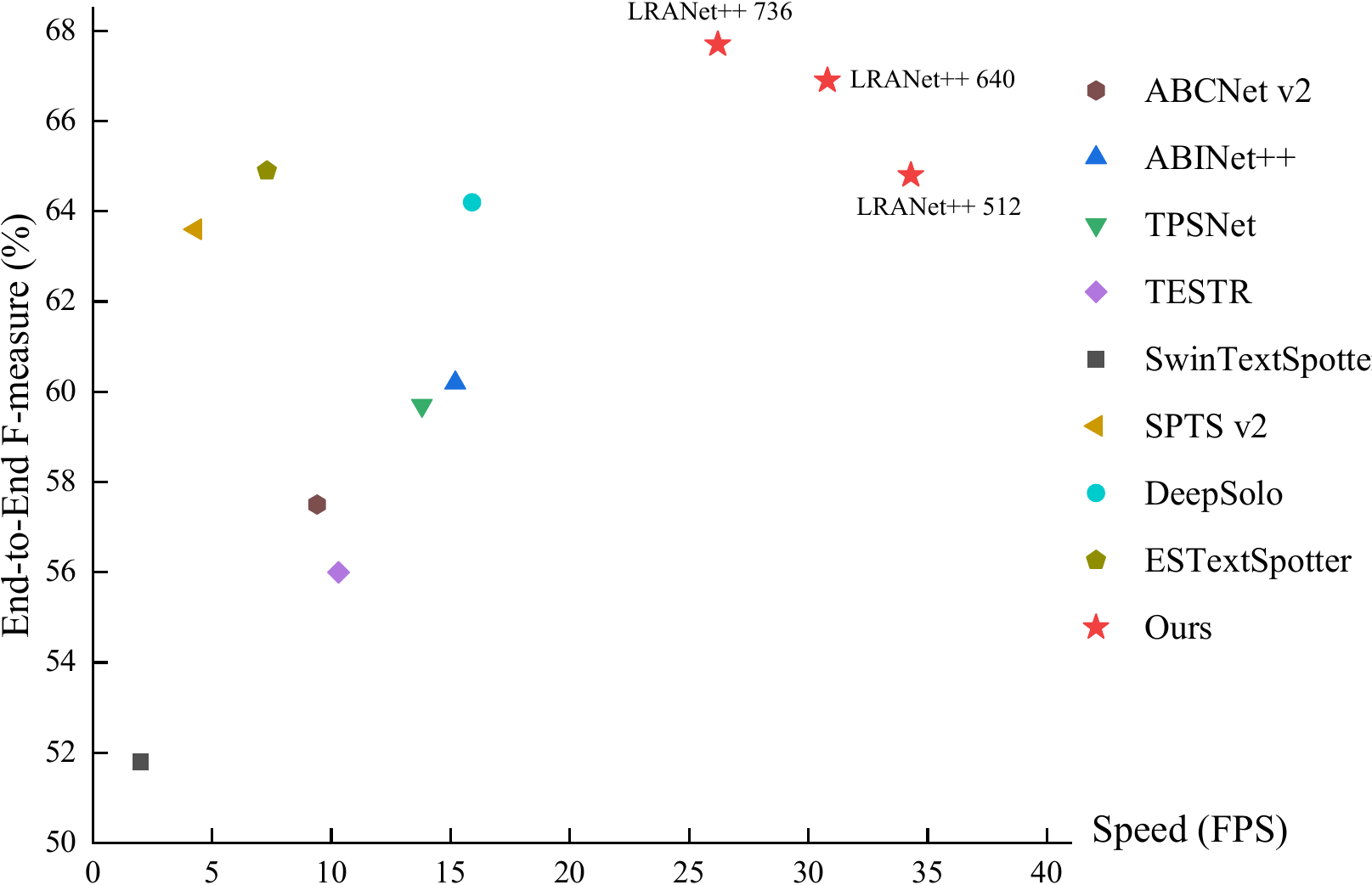}
    \caption{\textbf{The comparisons between our LRANet++ and several popular scene text spotters on  CTW1500 dataset.} LRANet++ achieves the leading F-measure while running much faster.}
    \label{fig-efficient-accurate}
\end{figure}

\begin{figure}[t]
    \centering
    \includegraphics[width=\linewidth]{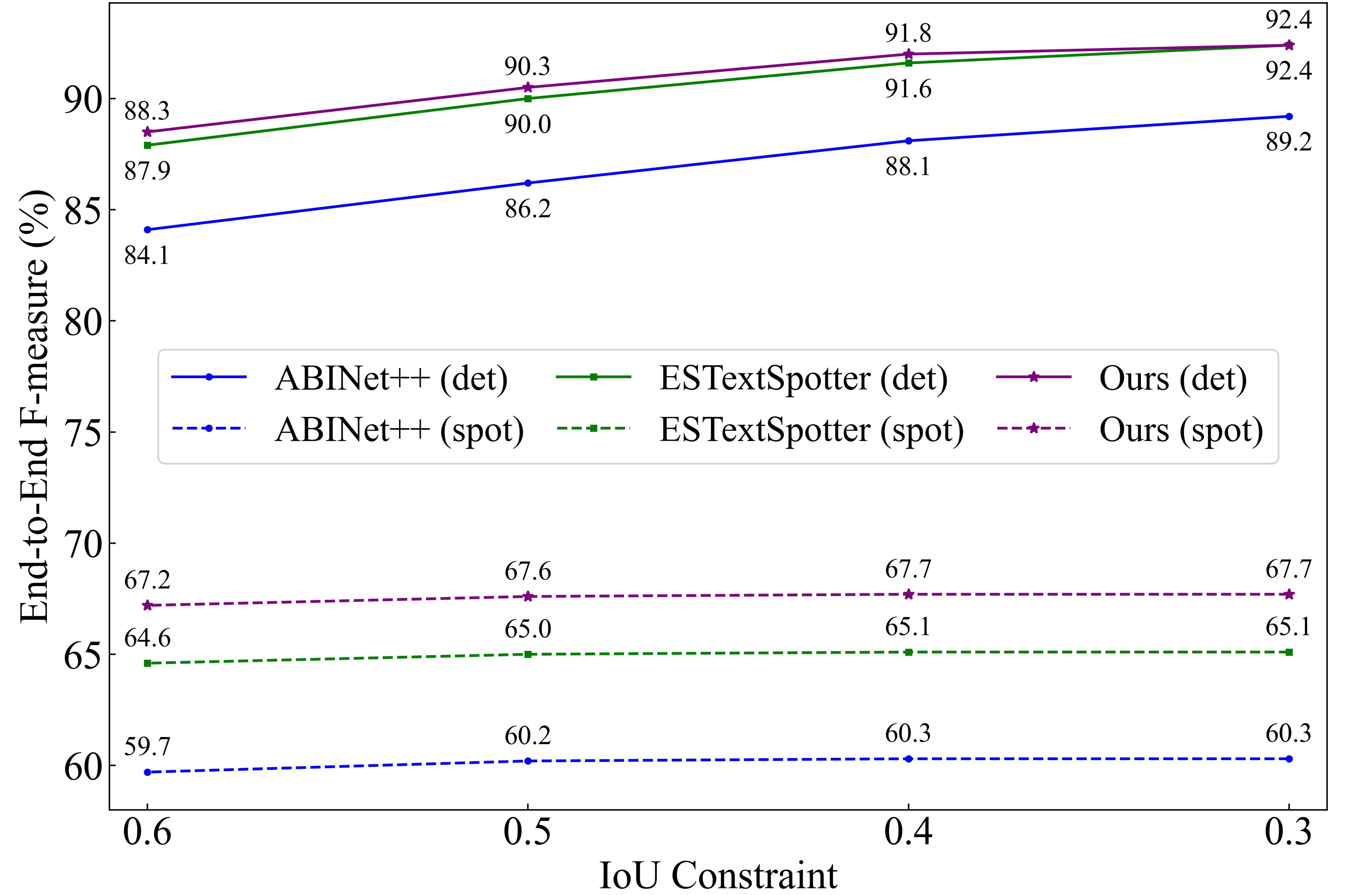}
    \caption{\textbf{Examples of F-measure variation under different IoU constraints.} It can be observed that inaccurate detection results (\textit{e.g.}, IoU with GT less than 0.5) rarely lead to accurate, and that fully capitalizing on well-localized text regions requires a well-designed overall spotting architecture.
    }\label{fig:ESTS_vs_ABI}
\end{figure}

We argue that the primary bottleneck for accurate and efficient text spotting lies in its inability to detect text precisely and quickly. Specifically, accurate localization is the prerequisite for extracting high-quality text content, while inaccuracies can lead to cumulative errors that impair recognition. Although some recent works have attempted to improve recognition from the perspective of reducing dependence on accurate localization. For example, ABINet++ \cite{fang2022abinet++} introduces a recognizer that incorporates language modeling to mitigate the impact of detection biases. ESTextSpotter \cite{huang2023estextspotter} proposes a task-aware decoder to model discriminative detection and recognition features in a decoupled manner. However, these methods have shown limited effectiveness in mitigating the cumulative errors caused by detection inaccuracies. As illustrated in Fig.~\ref{fig:ESTS_vs_ABI}, with the evaluation threshold of Intersection over Union (IoU) between detection results and ground-truth (GT) continuing to decrease, the F-measure of detection gradually increases, whereas the F-measure of spotting remains nearly unchanged. This indicates that inaccurate detection results (\textit{e.g.}, IoU with GT less than 0.5) rarely bring accurate spotting results. In fact, this finding aligns with the functioning of the human visual system, as without clear imaging on the retina, the brain's visual cortex cannot accurately recognize objects \cite{kandel2000principles}. Meanwhile, rapid detection is essential for fast text spotting. However, accurate and efficient detection methods have received less attention recently, while corresponding recognition techniques 
\cite{zhang2024self,du2025instruction,du2025context,du2025svtrv2} are flourishing. Consequently, we aim to build an accurate and efficient detection foundation for text spotting. Based on this foundation, we hope to achieve accurate and efficient text spotting by developing a well-designed overall spotting architecture that fully capitalizes on this strong detection.

To achieve this, we first analyze current detection methods, which can be roughly divided into segmentation-based methods \cite{zhao2024cbnet,zhang2022arbitrary,liao2022real} and regression-based methods \cite{wang2020textray,zhu2021fourier,su2024lranet}. The former models text instances with pixel-level classification masks that naturally fit arbitrary shapes, but they require costly post-processing to merge the results into text regions. More importantly, they mainly focus on local textual cues rather than the overall geometric layout of texts, leading to a lack of perception of text reading order. Thus, they are difficult to apply to spotting arbitrary-shaped text. In contrast, regression-based methods, which predict parameterized text shapes for text localization, are more suitable for spotting text, as they better consider the overall geometric layout of texts. In particular, they can implicitly learn the human reading order from ordered contour point annotations. However, there are still two main problems that limit their accuracy and efficiency.

One is that existing parameterized text shape methods still face challenges in modeling arbitrary-shaped text. Current methods \cite{liu2020abcnet,zhu2021fourier,wang2022tpsnet} mainly adopt contour points or parametric curves to fit the text shape. They either lack sufficient geometric constraints or fail to consider the distinct characteristics of text shapes, resulting in text boundaries not being faithfully represented. Specifically, scene text exhibits a wide range of shape diversity and aspect ratios. Current parameterized text shape methods solely model text shapes individually using data-irrelevant decomposition, ignoring the structural relationships among different text shapes and failing to exploit text-specific shape information. This makes it challenging to consistently and robustly represent various text shapes with only a few parameters.

\begin{figure}[t]
\centering
\includegraphics[width=\linewidth]{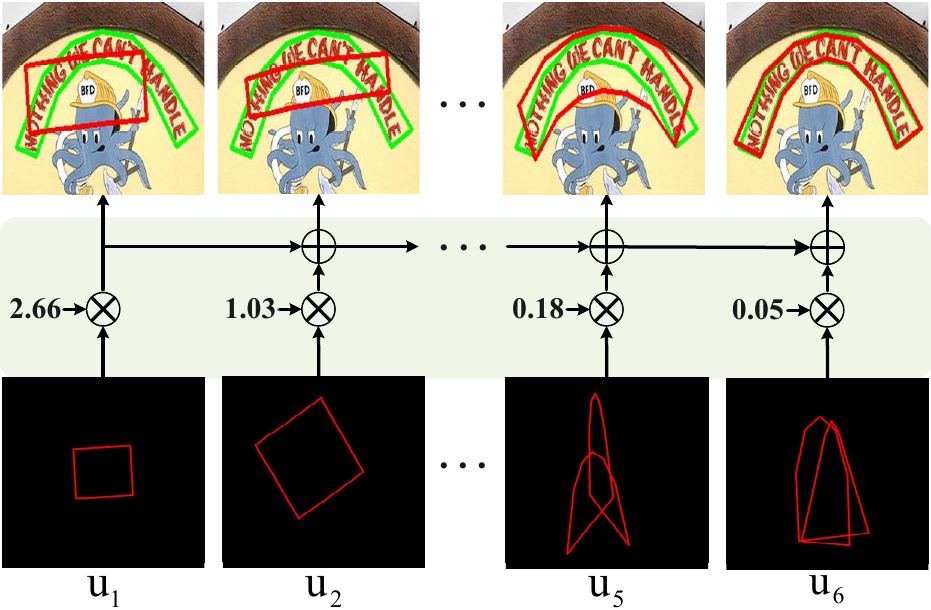}
\caption{\textbf{Illustration of the low-rank approximation representation.} The GT contour is depicted in green, with $u_1$,  $u_2$, ...,  $u_5$ and $u_6$ as $orthanchors$. The text contour is approximated by a linear combination of the $orthanchors$. As we can see, only six $orthanchors$ can fit the curved text.}
\label{fig-overview}
\end{figure}

The other is that regression-based methods often overlook the overall speed of the entire pipeline. Specifically, current regression-based methods can be categorized  into dense assignment approaches
\cite{liu2020abcnet,wang2022tpsnet} and one-to-one assignment approaches \cite{ ye2023dptext, shao2023ct} based on how positive samples are allocated. However, dense assignment approaches require non-maximum suppression (NMS) to filter a large number of redundant predictions, which is time-consuming for arbitrary-shaped text. Although one-to-one assignment approaches adopt the set prediction mechanism from DETR \cite{carion2020end} to mitigate this issue, they still lack sufficient supervised signals and explicit positional prior information. As a result, they usually stack multiple decoders for iterative text contour optimization, resulting in a complex pipeline.

Driven by the above analysis, we first propose a low-rank approximation (LRA) method to better represent arbitrary-shaped text. Unlike previous parameterized text shape methods that solely consider the individual text shape information, our LRA learns to represent text contours by exploring the shape correlation among different text instances. In detail, we first construct a text contour matrix, which contains all text contours in the training set. Then, we compute a low-rank subspace $\mathcal{S}$ to approximate the contour matrix. However, due to inevitable manual annotation errors, this contour matrix is inherently corrupted by outliers (noise). This makes classic methods like singular value decomposition (SVD) suboptimal, as its $\ell_2$-norm (squared error) objective is highly sensitive to these outliers, allowing large errors to be quadratically amplified and distort the computed subspace. Therefore, we adopt fast median subspace (FMS), a robust recovery method \cite{lerman2018fast} that minimizes absolute errors via its $\ell_1$-norm formulation. This $\ell_1$ formulation ensures the influence of large outliers is only linear, not quadratic, allowing FMS to recover a stable subspace not dominated by them. In this way, each text contour can be accurately represented by a very small number of orthogonal basis vectors, referred to as $orthanchors$. As illustrated in Fig.~\ref{fig-overview}, even extremely curved text can be faithfully reconstructed with a linear combination of as few as 6 vectors.

Next, we propose a triple assignment detection head. This design adopts a novel self-distillation architecture that decouples the inference path from complex learning tasks, effectively addressing the speed-accuracy compromise of prior designs \cite{su2024lranet}. Specifically, this head integrates three complementary paths: a dense branch provides rich supervision signals. Crucially, a deeper sparse branch learns to generate high-quality sparse positive samples, which are then used as the labels to teach the ultra-lightweight inference branch. This systematic design allows the inference branch to be significantly simplified, and by predicting sparse positive samples, it significantly reduces the NMS time for redundant predictions, thereby achieving an accurate and efficient inference of the $orthanchors$ coefficients.

Building upon these designs, we develop an accurate and efficient scene text detection foundation, which can replace the detection modules of existing text spotters (\textit{e.g.}, ABCNet v2 \cite{liu2021abcnet}, TPSNet \cite{wang2022tpsnet}), significantly improving their accuracy and efficiency. Finally, we integrate this detection foundation with a lightweight recognition branch to form an end-to-end text spotting method, termed LRANet++. This combination also needs careful design to fully harness its strong detection capabilities, as the final end-to-end performance can differ significantly even with comparable detection results, as shown in Fig.~\ref{fig:ESTS_vs_ABI}. Specifically, we design a lightweight yet accurate Transformer-based recognition head that adopts a progressive architecture to efficiently model global semantic context. Additionally, we retain traditional Region-of-Interest (RoI) operations for sampling recognition features. In contrast to the more recently popular dynamic sampling strategies \cite{ye2023deepsolo,huang2023estextspotter}, RoI operations preserve the modularity of text spotting, which can further accelerate inference in a producer-consumer pipeline. To address potential text distortion issues from RoI, we employ large-ratio image scaling data augmentation, which encourages the model to learn and adapt to the diverse deformations of text under various conditions. The performance advantages of our LRANet++ are shown in Fig.~\ref{fig-efficient-accurate}. The main contributions of this paper are summarized as follows:

\begin{enumerate}
\item We propose LRA, a novel parameterized text shape method. It represents text shapes faithfully by establishing a robust data-driven low-rank subspace.

\item We introduce a triple assignment scheme that decouples the learning complexity from the inference speed using a self-distillation framework.

\item Building upon the two contributions above, we design a new end-to-end arbitrary-shaped text spotting method, termed LRANet++. It achieves appealing trade-off between accuracy and inference speed. 

\item Extensive experiments show that LRANet++ gets state-of-the-art performance. In particular, LRANet++ is the first model to exceed 70\% in the end-to-end F-measure on CTW1500 to date, while achieving 26.2 FPS, which is 3.5x faster than the previous best method, LSGSpotter \cite{lyu2024arbitrary}.

\end{enumerate}

This paper is an extension of our previous work \cite{su2024lranet} that was accepted to AAAI’2024 as an oral presentation. Compared with its conference version, this paper introduces several new contributions, as outlined below:
\begin{enumerate}
\item We introduce FMS, a robust subspace recovery approach instead of SVD to compute a low-rank subspace more stably.
\item We extend the dual assignment detection head to a triple assignment detection head to further speed up inference.
\item We incorporate a lightweight recognition branch to develop an accurate and efficient text spotter.
\item We provide detailed methodological analyses and rich experimental validation for the design of our accurate and efficient text spotting method. 
\end{enumerate}

\section{Related Work}
\label{sec:related-works}

Scene text spotting aims to simultaneously detect and recognize text. It has evolved from the tasks of text detection and text recognition. In this section, we provide a brief review of these three tasks.

\subsection{Scene text detection}

\subsubsection{Segmentation-based Text Detection}

Segmentation-based methods \cite{liao2020Real, han2024real} treat text detection as a bottom-up segmentation problem. They first model text instances with pixel-level classification masks or character-level text components, and then combine them into text boundaries through specific heuristic operations. For example, 
DB \cite{liao2020Real} and its improved version DB++ \cite{liao2022real} introduce a differentiable binarization module that assigns a higher threshold to text boundaries, thereby allowing for distinction between adjacent text instances. TextPMs \cite{zhang2022arbitrary} proposes an iterative model to predict a set of probability maps, which are then grouped into text instances using region growth algorithms. SMNet \cite{han2024real} introduces a feature correction module that guides the model to suppress false positive predictions during the intermediate process. CBNet \cite{zhao2024cbnet} proposes a context-aware module to enhance text kernel segmentation and a boundary-guided module to adaptively expand the enhanced text kernel.

Despite progress, these methods lack global context awareness of text instances. This leads to sensitivity to background noise and an inability to infer reading order, making them less suitable for arbitrary-order text spotting.

\subsubsection{Regression-Based Text Detection}

Regression-based methods \cite{liu2020abcnet,wang2022tpsnet} are mainly inspired by general object detection, where text shapes are represented as vectors through parameterization methods for regression. Earlier approaches directly regress contour points to define the text boundary, but they fail to utilize prior information about its continuity. Therefore, later approaches use parameterized curves or parameterized masks to represent the text boundary. For example, TextRay \cite{wang2020textray} utilizes Chebyshev polynomials in the polar coordinate system to approximate text boundary. ABCNet \cite{liu2020abcnet} adopts the Bernstein polynomial to transform the long sides of the text into Bezier curves. FCENet \cite{zhu2021fourier} converts text contour points into Fourier signature vectors through Fourier contour embedding. TextDCT \cite{su2022textdct} transforms the text instance masks into the frequency domain by discrete cosine transform (DCT), and then extracts the low frequency components to approximate the text instance masks. TPSNet \cite{wang2022tpsnet} utilizes thin plate splines (TPS) to parameterize text contour points as TPS fiducial points.

However, these methods inadequately account for text-specific shape information, leading to limitations in representing arbitrary-shaped text. For example, Chebyshev polynomials and DCT representation struggle to accurately fit the text shape with a compact vector. Fourier representation may lose corner pixel information in long-text instances. Although TPS and Bezier polynomials can fit long-curved text through fiducial points, slight perturbations of these points induce significant shape distortions. Moreover, a limited number of fiducial points may fail to precisely represent arbitrary-shaped text. Unlike these methods, we propose LRA that represents the fitted curve from a low-rank basis vector decomposition perspective, allowing for effective utilization of text-specific shape information.

\subsection{Scene Text Recognition}

Scene text recognition \cite{du2022svtr,luo2025joint,li2026comprehensive} typically involves extracting visual features using a backbone network and aligning those features with their corresponding text sequences via a sequence-to-sequence (S2S) decoder. S2S can be categorized into two primary types: Connectionist Temporal Classification (CTC)-based and attention-based. CTC-based decoder \cite{du2022svtr, zhang2024self} aims to maximize the probability of all possible alignment paths that match the GT. It introduces blank labels and duplicate removal post-processing to address the alignment issue. (2) Attention-based decoder \cite{fang2021read,zheng2024cdistnet} utilize learnable queries and cross-attention operations to decode the recognition result in an autoregressive or parallel manner, with some recent works further integrating language information into the character decoding process to enhance recognition. For example, ABINet \cite{fang2021read} adopts an iterative refinement scheme, where linguistic knowledge is used to progressively correct recognition results with a standalone language model. PARSeq \cite{bautista2022scene} implicitly utilizes linguistic knowledge through a permuted auto-regressive (AR) sequence model for text recognition. VL-Reader \cite{zhong2024vl} utilizes masked vision and language models for auto-encoding and reconstruction to decode text, forcing effective cooperation between the visual and linguistic modalities.

In general, the CTC-based method has a much faster inference time but lower accuracy compared to the attention-based method. Thus, 
Zhang \etal~\cite{zhang2024self} introduce a framewise regularization term in the CTC loss to enhance individual supervision, and leverages maximum a posteriori estimation of latent alignment to resolve the inconsistency problem in distillation between CTC-based models. 
SVTRv2 \cite{du2025svtrv2} proposes a semantic guidance module to guide the CTC-based model to learn to perceive the linguistic context, achieving stunning performance in both speed and accuracy. However, the development of corresponding accurate and efficient text detectors is slow, hindering the advancement of effective text spotters.

\subsection{Scene Text Spotter}
\label{subsec:related-works-spotter}

Early text spotting methods \cite{jaderberg2016reading,liao2018textboxes++} simply connect independent detection and recognition models. Subsequent studies \cite{ronen2022glass,wang2022tpsnet,zhang2024inverse} show that jointly training both components can enhance overall performance. For instance, 
FOTS \cite{liu2018fots} introduces a RoI-Rotated operation to connect an oriented text detector with a text recognition module. To spot arbitrary-shaped scene text, PAN++ \cite{wang2021pan++} and MaskTextSpotter v3 \cite{liao2020mask} adopt RoI-Mask to filter background features with binary masks. ABCNet series \cite{liu2020abcnet, liu2021abcnet} propose the BezierAlign module to convert arbitrary-shaped text features into a horizontal representation. Similarly, TPSNet \cite{wang2022tpsnet} utilizes thin plate splines to align the detected features into a horizontal layout. To better spot inverse-like text, IAST \cite{zhang2024inverse} introduces a reading-order estimation module that learns reading-order information from text boundaries.

\begin{figure*}
\centering
\includegraphics[width=\textwidth]{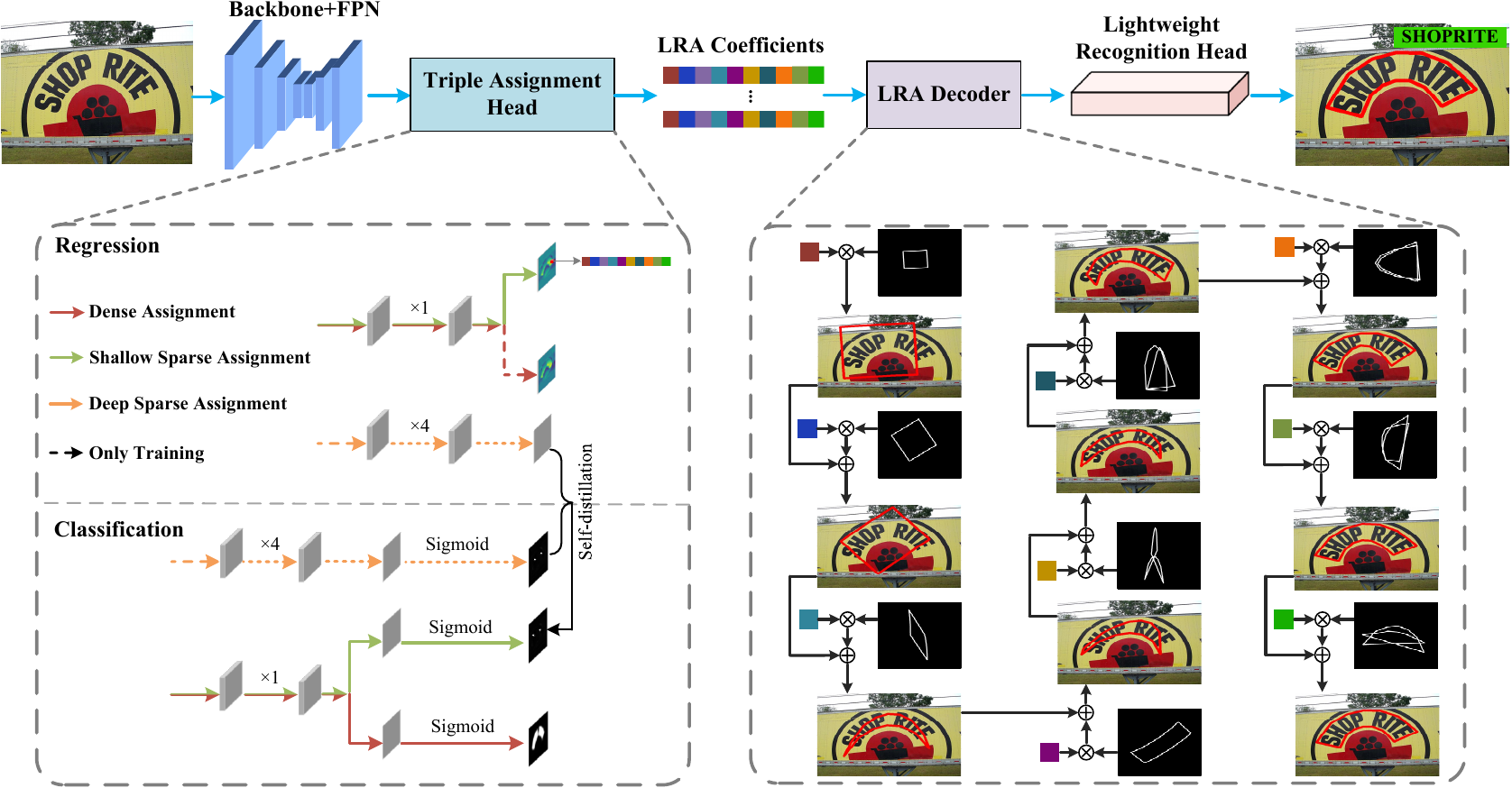}
\caption{\textbf{The overview of our LRANet++}. It is mainly composed of four modules: (a) backbone and FPN for feature extraction, (b) triple assignment head for predicting LRA coefficients, (c) LRA decoder to reconstruct text shape, and (d) lightweight recognition head that transcribes internal features of the text instance after TPS alignment into text sequences.}
\label{fig:architecture}
\end{figure*}

Inspired by DETR  \cite{carion2020end} and Pix2Seq \cite{chenpix2seq}, some works \cite{ye2023deepsolo,huang2023estextspotter,wan2024omniparser,lyu2024arbitrary} have explored the Transformer framework without complex post-processing and RoI, aiming to reduce error accumulation caused by inaccurate detection. 
For instance, TESTR \cite{zhang2022text} adopts two parallel Transformer decoders with shared queries for detection and recognition, respectively. DeepSolo \cite{ye2023deepsolo} designs instance queries based on centerline features and integrates the detector and recognizer into a single decoder with parallel prediction. ESTextSpotter \cite{huang2023estextspotter} further proposes task-aware query initialization to decouple the shared queries into separate detection and recognition queries. Although these methods achieve parallel prediction for detection and recognition, they still belong to the detection-then-recognition paradigm, as the extraction of recognition features relies on the positional information of the reference points generated by queries, which are directly related to detection. As shown in Fig.~\ref{fig:ESTS_vs_ABI}, inaccurate detection results rarely lead to accurate recognition, even though ESTextSpotter \cite{huang2023estextspotter} adopts a decoupled query scheme and ABINet++ \cite{fang2022abinet++} incorporates a language model to mitigate visual bias. Therefore, accurate detection is crucial for text spotting.

Moreover, although current text spotting methods mainly employ either a dynamic reference point sampling strategy \cite{huang2023estextspotter,ye2023deepsolo} or an auto-regressive image-to-sequence paradigm \cite{peng2022spts,wan2024omniparser,lyu2024arbitrary} to extract recognition features rather than using RoI operations (\textit{e.g.}, BezierAlign \cite{liu2020abcnet}, TPSAlign \cite{wang2022tpsnet}), this does not mean that RoIs are obsolete. This is because RoI-based fixed-size feature extraction ensures the modularity of the text spotting model, making it more suitable for real-time spotting of large batches in a producer-consumer pipeline. Specifically, during inference, the input image size often needs to be dynamically adjusted to ensure detection accuracy, but recognition can proceed in large batches due to the fixed-size features from RoI, thus accelerating inference without sacrificing accuracy. Admittedly, RoIs face two main challenges: first, the cumulative error caused by inaccurate detection; and second, text distortion caused by uniform feature size. For the first issue, as analyzed above, this cumulative error cannot be avoided, even with dynamic sampling methods. Moreover, minor detection errors may not impact recognition in RoI-based methods, as the extracted features cover larger receptive fields. To address the second issue, we apply large-ratio image scaling augmentation to enforce the model’s adaptation to diverse text deformations, mitigating distortion caused by fixed-size constraints. In summary, our LRANet++ still uses traditional RoI operations to extract features for recognition.

\section{Methodology}
\label{sec:method}

\subsection{Overview}
For effective and efficient text spotting, we adopt a compact single-shot fully convolutional architecture for text detection and follow a lightweight recognition head for text recognition. As shown in Fig.~\ref{fig:architecture}, our LRANet++ mainly comprises four parts: a feature extraction module, a detection head, an LRA decoder, and a recognition head. Specifically, in the feature extraction module, we utilize ResNet50 \cite{he2016deep} with DCN \cite{zhu2019deformable} as our backbone network, and adopt the Feature Pyramid Network (FPN) \cite{lin2017feature} to extract multi-scale features. These features are then fed into the triple assignment detection head to predict LRA coefficients. Subsequently, the LRA decoder transforms the predicted LRA coefficients into the text contours. Finally, the lightweight recognition head transcribes internal features of the text instance into text sequences after the RoI operation.

\subsection{Low-Rank Approximation Representation}

LRA is a widely used  technique for dimensionality reduction. In this paper, we first introduce LRA to compactly represent text contours. Unlike previous parameterized text shape methods that use curve fitting \cite{liu2020abcnet,zhu2021fourier,wang2022tpsnet} or mask compression \cite{su2022textdct}, LRA is a data-driven approach that represents text boundaries in a low-dimensional space by exploiting the distribution of labeled text boundaries.

Scene text shapes are typically well-structured, mainly characterized by large aspect ratios and right-angle corners. As a result, there is significant correlation among these text shapes. By exploiting this correlation based on labeled data, we design a novel parameterized text shape method. Specifically, the GT text boundary typically consists of multiple vertices, we first flatten them into a column vector $\mathbf{p}=\left[\mathbf{x}_1, \mathbf{y}_1, \cdots, \mathbf{x}_N,\mathbf{y}_N\right]^{\top}\in{\mathbb{R}^{{2N} \times 1}}$, where $N$ is the number of vertices. Then, we construct a text contour matrix $\mathbf{A}=\left[\mathbf{p}_1, \mathbf{p}_2, \cdots, \mathbf{p}_L\right]\in{\mathbb{R}^{{2N} \times L}}$ from a set of $L$ labeled text instances. Finally, we use low-rank subspace projection to effectively capture the structural relationships among the text contours in matrix $\mathbf{A}$, and employ low-rank reconstruction to approximate the text contours.

\subsubsection{Low-Rank Subspace Projection}

SVD is the classical and most commonly used method for computing a low-rank subspace $\mathcal{S}$  with dimension $M\ll2N$ from the text contour matrix $\mathbf{A}\in{\mathbb{R}^{{2N} \times L}}$. 
However, this approach is suboptimal because the contour matrix $\mathbf{A}$ is inevitably corrupted by manual annotation errors (outliers). SVD's limitation stems from its $\ell_2$-norm (squared error) objective, which is non-robust as large errors from these outliers are quadratically amplified, potentially distorting the computed subspace.

Therefore, to robustly estimate the underlying low-dimensional subspace of matrix $\mathbf{A}$ in the presence of outliers, we utilize FMS \cite{lerman2018fast},  which efficiently computes a basis $\mathbf{U}_M\in\mathbb{R}^{2N\times M}$ for the subspace $\mathcal{S}$  by solving the non-convex least absolute loss problem:
\begin{align}\label{eq:fms}
\mathbf{U}_M = \arg\min_{\mathbf{U} \in \mathcal{O}(2N,M)} \sum_{j=1}^{L} \left\| (\mathbf{I} - \mathbf{U}\mathbf{U}^\top ) \mathbf{p}_j \right\|_2\,.
\end{align}
Here, $\mathcal{O}(2N,M):=\{\bm U\in\mathbb{R}^{2N\times M}: \bm U^\top \bm U  = \bm I_{M}\}$ denotes the set of orthonormal matrices. It aims to minimize the sum of Euclidean distances from all text contours in matrix $\mathbf{A}$ to their projections onto $\mathcal{S}$. Geometrically, FMS estimates a ``median" basis for the underlying $M$-dimensional subspace, which is more robust to outliers than the traditional ``mean" basis estimation.

Since the $\ell_{1,2}$-norm objective in Eq. (\ref{eq:fms}) is non-smooth, it cannot be solved with a single decomposition like SVD. Instead, FMS computes the basis $\mathbf{U}_M$ through an iterative process known as Iteratively Reweighted Least Squares (IRLS) \cite{daubechies2010iteratively}, which we summarize as in Algorithm \ref{alg:fms_computation}.

Thus, the subspace projection $\mathbf{C}_M$ of matrix $\mathbf{A}$ is computed as: 
\begin{align}\label{eq:UA}
\mathbf{C}_M = {\mathbf{U}_M}^\top \mathbf{A} = \left[\mathbf{c}_1, \mathbf{c}_2, \cdots, \mathbf{c}_{L}\right] \quad \in{\mathbb{R}^{{M} \times L}}\,,
\end{align}
where the subspace basis $\mathbf{U}_M$ captures the most significant structural components of matrix $\mathbf{A}$, and  $\mathbf{c}_i \in{\mathbb{R}^{{M} \times 1}}$ represents the projection of the contour $\mathbf{p}_i$ onto the subspace.

\begin{algorithm}[t]
\caption{Fast Median Subspace (FMS) Projection}
\label{alg:fms_computation}
\begin{algorithmic}[1] 
    \STATE \textbf{Input:} Text contour matrix $\mathbf{A} \in \mathbb{R}^{2N \times L}$,
    Subspace dimension $M$, Max iterations $T_{max}$, Tolerance $\tau$
    \STATE \textbf{Output:} The robust $M$-dimensional subspace basis $\mathbf{U}_M$
    
    \STATE \textbf{Initialize:} $k \leftarrow 0$.
    \STATE $\mathbf{U}^{(0)} \leftarrow \text{SVD}(\mathbf{A}, M)$  \COMMENT{Initialize with standard SVD}
    
    \REPEAT
        \STATE $k \leftarrow k+1$
        \STATE $\mathbf{W}^{(k)} \leftarrow \mathbf{I}_{L \times L}$ \COMMENT{Initialize diagonal weight matrix}
        
        \FOR{$j = 1$ to $L$}
            \STATE $\mathbf{r}_j^{(k-1)} \leftarrow (\mathbf{I} - \mathbf{U}^{(k-1)} (\mathbf{U}^{(k-1)})^\top ) \mathbf{p}_j$ \COMMENT{Compute residual}
            \STATE $w_{j j} \leftarrow 1.0 / \max\left(\left\|\mathbf{r}_j\right\|_2, \epsilon\right)$ \COMMENT{Compute robust $\ell_1/\ell_2$ weight}
        \ENDFOR

    \STATE $\mathbf{U}^{(k)} \leftarrow \arg\min\limits_{\mathbf{U} \in \mathcal{O}(2N,M)} \sum_{j=1}^{L} w_{j j}^{(k)} \left\| (\mathbf{I} - \mathbf{U}\mathbf{U}^\top ) \mathbf{p}_j \right\|_2^2$
        
    \UNTIL{$k \ge T_{max}$ \textbf{or} $dist(\mathbf{U}^{(k)}, \mathbf{U}^{(k-1)}) < \tau$}
    
    \STATE \textbf{Return} $\mathbf{U}_M \leftarrow \mathbf{U}^{(k)}$
\end{algorithmic}
\end{algorithm}

\subsubsection{Low-Rank Reconstruction}  

To recover the approximation of the matrix $\mathbf{A}$, we perform the reconstruction by
multiplying the coefficient matrix \(\mathbf{C}_M\) with  the basis \(\mathbf{U}_M\), which maps the low-dimensional representation back to the original space:
\begin{align}\label{eq:UUA}
  \mathbf{A}_M =\mathbf{U}_M\mathbf{C}_M = \left[\tilde{\mathbf{p}}_1, \cdots, \tilde{\mathbf{p}}_L\right] \approx \mathbf{A}\,,
\end{align}
where $\tilde{\mathbf{p}}_i$ denotes the approximation of $\mathbf{p}_i$. In other words,
\begin{equation}
\label{eq-p=uc}
\tilde{\mathbf{p}}_i=\mathbf{U}_M \mathbf{c}_i=\left[\mathbf{u}_1, \cdots, \mathbf{u}_M\right] \mathbf{c}_i\,.
\end{equation}
We call these $\mathbf{u}_1, \cdots, \mathbf{u}_M$
as $orthanchors$, as they are a set of orthonormal basis vectors and can be viewed as pre-defined arbitrary-shaped anchors, as shown in Fig.~\ref{fig-overview}.

We refer to the space spanned by the  $orthanchors$ as  $orthanchor$ $space$. 
For any $2N$-dimensional text contour $\mathbf{p}$, the approximation can be reconstructed by Eq.~\eqref{eq-p=uc}, since $\mathbf{U}_M$ is fixed after solving Eq.~\eqref{eq:fms}, only the low-dimensional coefficient vector $\mathbf{c}_i$ needs to be predicted to approximate the contours. Note that the number of contour vertices in $\mathbf{p}$ may differ from $N$. In such case, we resample $N$ vertices from $\mathbf{p}$ using cubic spline interpolation.

\subsection{Triple Assignment Detection Head}

To accurately and efficiently regress LRA coefficients, an ideal detection head should satisfy several challenging requirements. First, for efficient post-processing, it must generate sparse predictions to minimize the time consumed by redundant filtering (such as NMS). Second, for efficient inference, the head architecture itself must be lightweight, as it is typically replicated across multiple FPN feature layers (such as P3, P4, and P5) to regress shape parameters at different scales, meaning any architectural complexity is significantly amplified. Third, to ensure training accuracy, this sparse predictor requires high-quality, representative positive samples (\emph{i.e.}, a ``precise'' signal). Fourth, the head simultaneously requires ample dense supervision (\emph{i.e.}, a ``broad'' signal) to effectively learn robust features.

We propose a triple assignment head to resolve these requirements. It uses an asymmetric ``Teacher-Student-Auxiliary'' architecture to decouple the complex training task from the lightweight inference task. As illustrated in Fig.~\ref{fig:architecture}, the ``Student'' is an ultra-lightweight, shallow sparse assignment branch that is used alone for inference. It is guided during training by two ``training-only'' branches: the ``Teacher'', a deep sparse assignment branch, provides the precise positive sample regions, while the ``Auxiliary'', a dense assignment branch, provides the broad supervision signals. Here, ``shallow'' and ``deep'' refer to the number of convolutional layers within each branch, while ``sparse'' and ``dense'' denote the number of positive samples assigned.

Specifically, in the deep sparse assignment branch (``Teacher''), we construct a prediction-aware matrix $\mathbf{S}\in{\mathbb{R}^{{HW} \times KT}}$ for selecting $K$ positive samples for each text instance, where $H$ and $W$ are the height and width of output features, and $T$ denotes the number of text instances in each image. The matrix element is defined as: 
\begin{equation}
\label{eq-Selection-metric}
{s}_{i j}=\left\{\begin{array}{cc}
 \mathrm{FL}^{\prime}({b}_{i}) + \lambda \sum\limits_{n=0}^{N-1}\left\|{\tilde{\mathbf{p}}}_{\raisebox{-0.2ex}{$\scriptstyle i$}}^{(n)}-\mathbf{p}_{\raisebox{-0.2ex}{$\scriptstyle j$}}^{(n)}\right\|,    & i \in TR \\[3ex]
 \infty, & i \notin TR
\end{array} . \right .
\end{equation}
Here, ${s}_{i, j}$ denotes the matching cost between the $i$-th point and the $j$-th GT text instance, ${b}_{i}$ is the predicted classification score of the $i$-th point. $\mathrm{FL}^{\prime}$ is defined as the difference between the positive and negative terms: $\mathrm{FL}^{\prime} = -\alpha(1-x)^\gamma \log (x)+(1-\alpha) x^\gamma \log (1-x)$, which is derived from the Focal loss \cite{lin2017focal}. We set $\alpha$ to $0.25$ and $\gamma$ to $2.0$. The second term is the $\ell_1$ distance between the $i$-th predicted contour and $j$-th GT contour, and $\lambda$ controls the importance degree of classification and regression. The third item aims to limit the sparse sampling to only the text region for better joint optimization. 

Afterwards, we regard the sparse positive sampling as a bipartite matching problem and use the Hungarian algorithm to solve the matrix $\mathbf{S}$ in ascending order, to find the optimal matching point for each text instance. To explore the optimal number of positive sample allocations, we replicate it $K-1$ times when constructing the matrix $\mathbf{S}$, and thus assign $K$ positive samples to each instance. However, this learning introduces an issue where its training labels are dynamically generated from its own prediction outputs. This self-referential loop creates an interdependent relationship that necessitates a deep branch structure to effectively learn and stabilize the process. Thus, we utilize four $3 \times 3$ convolutional layers with 256 output channels to extract task-specific features for the ``Teacher'' branch.

In our shallow sparse assignment branch (``Student''), we treat the positive sample regions computed by the deep sparse assignment branch as the GT. Since the classification learning objective is a simple binary classification task and regression and classification are highly correlated, this branch can be implemented with a very concise structure. Thus, we adopt a single $3 \times 3$ convolutional layer with 32 output channels to extract task-specific features for this branch. Meanwhile, the dense assignment branch (``Auxiliary'') shares this structure and treats the text region as the positive sample region.

In summary, the dense assignment branch and the deep sparse assignment branch assist in training the shallow sparse assignment branch by providing abundant supervised signals and accurate positive sample regions, respectively. During inference, we utilize the sparse positive samples predicted by the shallow sparse assignment branch and its lightweight structure to accelerate post-processing and model inference, respectively.

\subsection{Recognition Head}

\begin{figure}[t]
    \centering
    \includegraphics[width=\linewidth]{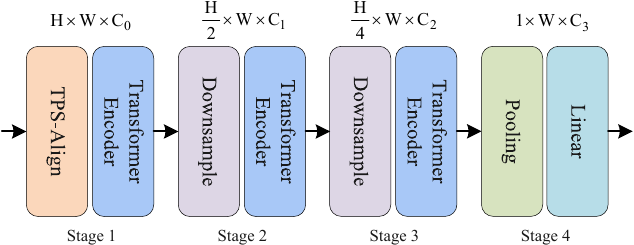}
    \caption{\textbf{The structural details of the recognition head.} It comprises a four-stage network with progressively decreasing height, and recognition is ultimately performed through a linear prediction layer.}
    \label{fig:recog_head}
\end{figure}

For efficiency, we choose the CTC decoder over the autoregressive decoder in the recognition stage. However, current CTC decoder methods \cite{liu2020abcnet,ye2023deepsolo} either neglect global semantic information modeling, resulting in a lack of accuracy, or have redundant structures, leading to inefficiency. To address this, we design a Transformer-based recognition head consisting of a four-stage network with progressively decreasing height, as shown in Fig.~\ref{fig:recog_head}. 

First, TPS alignment is applied to sample features from the text regions of the FPN. Then, in the first three stages, we use $L$ Transformer encoder layers to model the global semantic information of the text features. In this process, we do not introduce any biases (such as local windows), aiming for adaptive learning of regions of interest, allowing the model to fully leverage the power of Transformers in processing large-scale data. Next, maintaining a constant spatial resolution across stages results in high computational cost and redundant representations. Thus, we apply a $3 \times 3$ convolution with a stride of 2 along the height dimension at the beginning of stages 2 and 3, reducing the height of the feature map. Finally, in the final stage,  the height of the feature map is pooled to 1 to form a feature sequence suitable for text transcription, and recognition is performed using a simple linear prediction with the CTC decoder.

\subsection{Overall Loss}

In LRANet++, the overall loss is formulated as:
\begin{equation}
\label{loss_total}
\mathcal{L}=\mathcal{L}_{det}+ \mathcal{L}_{rec}, 
\end{equation}
where $\mathcal{L}_{det}$ and $\mathcal{L}_{rec}$ are the loss functions for text detection and recognition, respectively.

Specifically, the text detection loss $\mathcal{L}_{det}$ can be written as:
\begin{equation}
\label{loss_det}
\mathcal{L}_{det}=\mathcal{L}_{cls}+ \mathcal{L}_{reg}, 
\end{equation}
where $\mathcal{L}_{cls}$ and $\mathcal{L}_{reg}$ are the losses for foreground classification and LRA coefficient regression, respectively.
The classification loss $\mathcal{L}_{cls}$ 
is composed of the text region loss  $\mathcal{L}_{tr}$ and the sparse sampling region loss $\mathcal{L}_{ssr}$:
\begin{equation}
\label{loss2}
\mathcal{L}_{cls}=\lambda_{tr}\mathcal{L}_{tr}+ \lambda_{ssr}\mathcal{L}_{ssr},  
\end{equation}
where $\mathcal{L}_{tr}$ and $\mathcal{L}_{ssr}$ are the cross entropy loss and Focal loss, respectively. The regression loss is defined as:
\begin{equation}
\mathcal{L}_{reg}=\mathds{1}^{\mathcal{P}} \sum_{i}^{N_{\mathcal{P}}} \text{Smooth-}\ell_1\left(\tilde{\mathbf{p}}_{i}, \mathbf{p}_{i}\right),
\end{equation}
where $\mathcal{P}$ is the positive sample region in our triple assignment scheme, $\mathds{1}$ is a spatial indicator, outputting $1$ when the point is within $\mathcal{P}$ and $0$ otherwise.

For text recognition, we adopt the
CTC loss \cite{graves2006connectionist} for transcribing variable-length text:
\begin{equation}
\mathcal{L}_{rec}= \text{CTC}\left(\tilde{\mathbf{t}}_{i}, \mathbf{t}_{i}\right),
\end{equation}
where $\mathbf{t}_{i}$ and $\tilde{\mathbf{t}}$
denote the predicted and GT text sequences, respectively.

\section{Experiments}\label{sec:exp}

\subsection{Datasets} 
    
\textbf{Synth150K} \cite{liu2020abcnet} is a synthetic dataset consisting of $54,327$ curved text images and
$94,723$ multi-oriented images.
\textbf{Total-Text} \cite{ch2017total} is an arbitrary-shaped word-level scene text dataset. It contains
 $1,255$ images, with $1,000$ for training and $255$ for testing. Each image contains at least one example of curved text. \textbf{CTW1500} \cite{liu2019curved} is another important arbitrary-shaped scene text benchmark, containing $1,000$ training and $500$ test images. 
 \textbf{MSRA-TD500} \cite{yao2012detecting} is a multi-language text detection dataset that consists of $300$ training images and $200$ test images.
 \textbf{ICDAR 2013 (IC13)} \cite{karatzas2013icdar} is a horizontal dataset that contains $229$ training and $233$ test images.
 \textbf{ICDAR 2015 (IC15)} \cite{karatzas2015icdar} is a multi-oriented scene text dataset that includes $1,000$ training and $500$ test images. The text instances are labeled at the 
 word-level.
\textbf{ICDAR17 MLT (MLT17)} \cite{nayef2017icdar2017} is a multi-language scene text dataset containing $9,000$ training images. 
\textbf{TextOCR} \cite{singh2021textocr} is currently the largest real dataset for text spotting. It is composed of $21,749$ training, $3,153$ validation, and $3,232$ test images. \textbf{Inverse-Text} \cite{ye2023dptext} is a recently proposed dataset focused on inverse-like scenes. It consists of $500$ images for testing, with about 40\% being inverse-like instances. \textbf{ICDAR19 ArT} \cite{chng2019icdar2019} is currently the largest arbitrary-shaped dataset. It contains $5,603$ training images. \textbf{ICDAR19 LSVT} \cite{sun2019icdar} is a large-scale Chinese scene text dataset with 30,000 training images. \textbf{ReCTS} \cite{zhang2019icdar} is a benchmark for Chinese text on signboards, containing 20,000 training and 5,000 test images. \textbf{VinText} \cite{nguyen2021dictionary} is a Vietnamese text dataset, including 1,200 training and 500 testing images.

\begin{figure}[t]
\centering
\includegraphics[width=\linewidth]{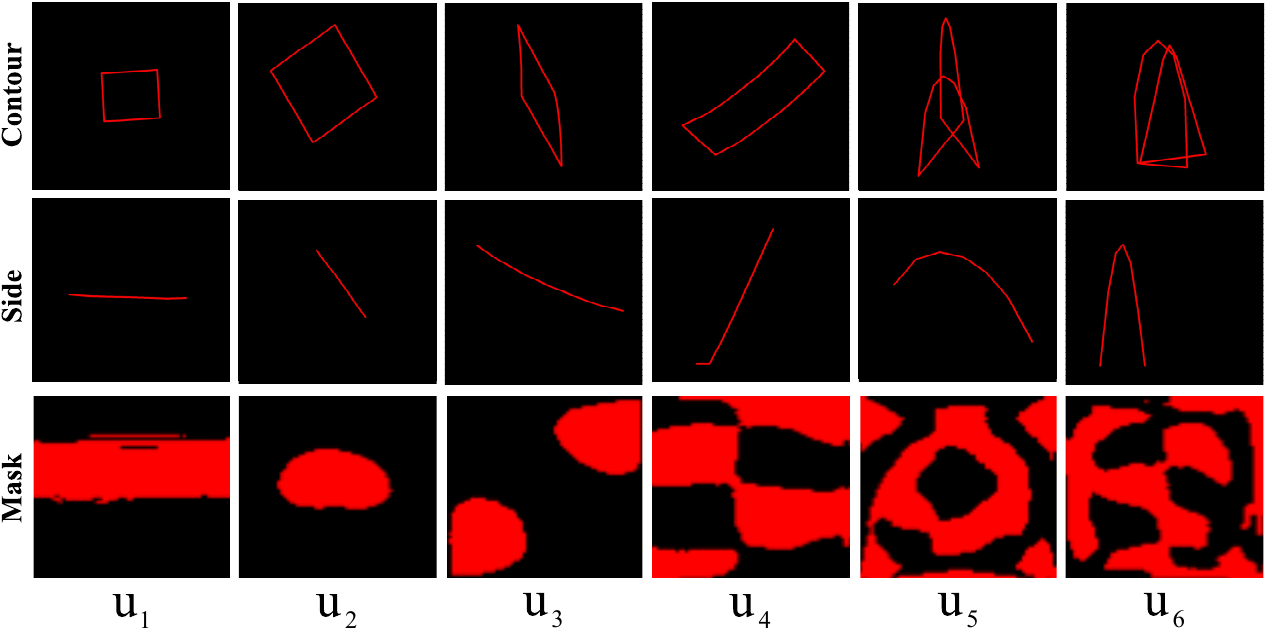}
\caption{\textbf{Visualization of the first six orthanchors with different data-driven.} These $orthanchors$ are obtained from the CTW1500 training dataset. 
}
\label{fig:mask-line-polygon}
\end{figure}

\begin{table}[!t]
\caption{\textbf{Comparison of different data representation on CTW1500.} IoU refers to the intersection over union between reconstructed text region and GT text region. E2E: End-to-end}
\label{tab:mask-side-contour}
 \setlength\tabcolsep{9pt}
    \centering
    \begin{tabular}{c|ccccc}
    \toprule
    \multirow{2}{*}{Data-Driven} & \multicolumn{3}{c}{Detection} & 
    \multicolumn{1}{c}{E2E} & \multirow{2}{*}{IoU}\\ 
\cmidrule(l{0.5em}r{0.5em}){2-4} 
\cmidrule(l{0.2em}r{0.2em}){5-5}
     & R & P & F & F  & \\ 
    \midrule
    Mask & 84.9 & 87.3 & 86.1 & 65.1  & 90.9 \\
    Side & 85.2 & 88.9 & 87.0 & 67.3 & 97.9 \\
    Contour  & \textbf{85.3} & \textbf{89.1} & \textbf{87.2} & \textbf{67.7} & \textbf{98.1} \\
    \bottomrule
    \end{tabular}%
\end{table}

\begin{figure}[t]
    \centering
    \includegraphics[width=\linewidth]{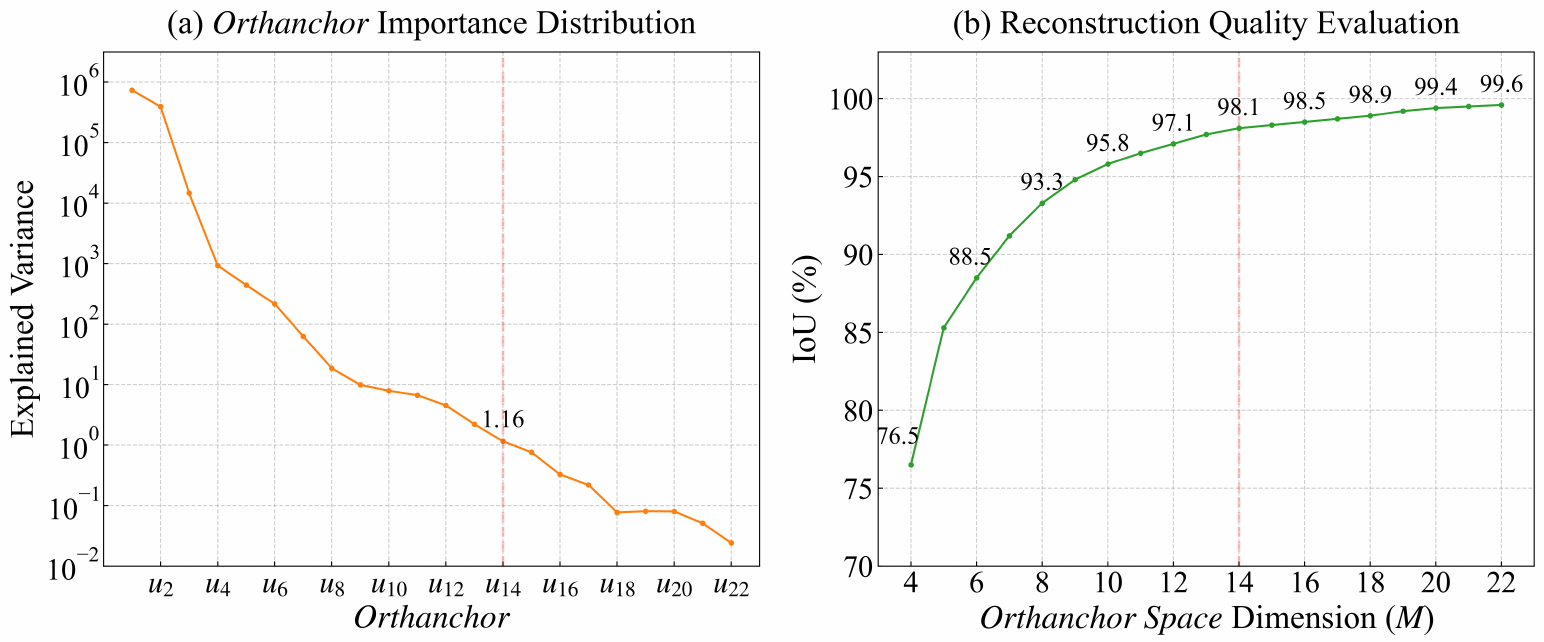}
    \caption{\textbf{Analysis of the FMS-derived basis vectors (termed $orthanchors$).} (a) \textbf{Importance evaluation:} Explained variance for each orthanchor. (b) \textbf{Reconstruction quality evaluation:} Intersection over Union (IoU) between the reconstructed and GT regions.}
    \label{fig:Analysis_U}
\end{figure}

\subsection{Implemented Details}
\label{subsec:imd}

The dimension $M$ of the $orthanchors$ is set to $14$. The sampling number $K$ in the sparse assignment scheme is $3$. The number of Transformer encoder layers 
$L$ in the recognition head is 3. The loss weight $\lambda$ in Eq.~\eqref{eq-Selection-metric} is $2$, while the loss weights $\lambda_{tr}$ and $\lambda_{ssr}$ in Eq.~\eqref{loss2} are set to $1$ and $2$, respectively. The feature map size after TPS alignment is $8\times64$ on CTW1500 and $8\times32$ on other datasets. The character type is 97, and the maximum length of the output text is 25. 
The employed data augmentation includes random rotation, random scaling, random crop, and color jitter.

In the text detection task, to make fair comparisons, we train LRANet++ (without the recognition head) with two strategies as follows: 1) train the model for 500 epochs on each real dataset without using external text datasets; 2) pre-train the model on Synth150K for $2$ epochs, and then fine-tune it for $300$ epochs on each real dataset.

In the text spotting task, we adopt different pre-training configurations for different languages. For English datasets, two main strategies are used: 1) Following the training strategy in \cite{liu2021abcnet,wang2022tpsnet,fang2022abinet++}, the model is pre-trained on the Synth150K, ICDAR17 MLT, and Total-Text datasets for $250k$ iterations; 2) Pre-training on a larger set of data, including Synth150K, TextOCR, ICDAR 2013, ICDAR 2015, ICDAR17 MLT, and Total-Text, for $450k$ iterations. For Chinese, following previous works \cite{fang2022abinet++,huang2023estextspotter,huang2025swintextspotter}, we adopt the Chinese synthetic pretrained data \cite{liu2021abcnet}, ICDAR19 ArT, ICDAR19 LSVT and ReCTS to pre-train the model for $250k$ iterations. For VinText, the training strategy is consistent with \cite{huang2023estextspotter, huang2025swintextspotter}.

\begin{table}[!t]
\caption{\textbf{Experimental results of different $orthanchor$ $space$ dimension on CTW1500.}}
\label{tab:lrsp2}
 \setlength\tabcolsep{12pt}
    \centering
    \begin{tabular}{c|ccccc}
    \toprule
    \multirow{2}{*}{Dim} & \multicolumn{3}{c}{Detection} & \multicolumn{1}{c}{E2E} & \multirow{2}{*}{IoU}\\ 
\cmidrule(l{0.5em}r{0.5em}){2-4} 
\cmidrule(l{0.2em}r{0.2em}){5-5}
    & R & P & F & F & \\ 
    \midrule
       10 & 84.9 & 88.3 & 86.6  & 67.1  & 95.8 \\
14 & 85.3 & \textbf{89.1} & \textbf{87.2}  & 67.7  & 98.1 \\ 
       18 & \textbf{85.6} & 88.5 & 87.1  & \textbf{67.9}  & \textbf{98.9} \\
    \bottomrule
    \end{tabular}%
\end{table}

In the testing stage, we set the short sides of the test images to $736$, $1000$, $1000$, $800$, $1440$, $1056$, $1056$ for CTW1500, Total-Text, Inverse-Text, MSRA-TD500, and ICDAR 2015, ReCTS, and VinText, respectively.
Following \cite{liu2021abcnet,ye2023dptext,wang2022tpsnet}, we adopt IoU@0.5 as the evaluation metric for text detection across all datasets. The evaluation metric for text spotting is consistent with \cite{liu2020abcnet,huang2022swintextspotter,huang2023estextspotter}. The model is trained distributively on 4 NVIDIA RTX 3090 GPUs with a batch size of 4 per GPU, and all the inference speeds listed in the tables are tested on a single NVIDIA RTX 3090 GPU. When using the $250k$ iteration pre-training schedule for text spotting, the whole training process takes about 2.5 days.

\subsection{Ablation Study} 
\label{subsec:abs}

To deeply analyze LRANet++, we conduct ablation studies on CTW1500 and Total-Text datasets. In these experiments, the text detection model is trained without pre-training on external text datasets, and we adopt the first pre-training dataset described in Sec. \ref{subsec:imd} for training the text spotting model, without using lexicons.

\subsubsection{Different Data Representations}

The current parameterized text shape methods can be categorized into three types: parameterized text contour \cite{zhu2021fourier}, parameterized text mask \cite{su2022textdct}, and parameterized top and bottom sides \cite{liu2020abcnet}. To investigate the impact of different data representations for LRA, we construct the matrix $\mathbf{A}$ based on each of these three types, setting $M$ to $14$, with the mask size being ($8\times64$) for the parameterized text mask method. Qualitative results are shown in Fig.~\ref{fig:mask-line-polygon}. The mask-driven method contains redundant information, making it difficult to extract text shapes accurately. In contrast, the side-driven and contour-driven methods can extract text shape information more accurately, with the contour-driven method excelling at capturing complex text shape information. Consistent with this observation, the results in Table~\ref{tab:mask-side-contour} demonstrate that the mask-driven method exhibits poor representation quality, resulting in worse detection performance compared to the other two methods. Furthermore, ROI-Mask \cite{wang2021pan++} is naturally suited to the mask-driven method to extract features for recognition. However, it shows significantly lower text spotting performance, with a 2.6\% reduction in the F-measure compared to the contour-driven method, which may be due to the ROI-Mask containing excessive background information. As a result, the contour-driven method achieves the highest representation quality and superior text detection and spotting performance.

\begin{table}[!t]
\caption{\textbf{Robustness comparison of FMS and SVD on CTW1500 under different ratios of artificial label noise.}}
\label{tab:noise}
 \setlength\tabcolsep{7pt}
    \centering
    \begin{tabular}{c|c|ccccc}
    \toprule
    \multirow{2}{*}{Noise} & \multirow{2}{*}{Method} & \multicolumn{3}{c}{Detection} & \multicolumn{1}{c}{E2E} & \multirow{2}{*}{IoU}\\ 
\cmidrule(l{0.5em}r{0.5em}){3-5} 
\cmidrule(l{0.2em}r{0.2em}){6-6}
    &  & R & P & F & F & \\ 
    \midrule \multirow{2}{*}{5\%}   &
    SVD & 84.9 & 88.1 & 86.5  & 67.1  & $95.0^{(\downarrow3.0)}$ \\
    & FMS & \textbf{85.5} & \textbf{88.5} & \textbf{87.0}  & \textbf{67.7}  & $\mathbf{97.6}^{(\downarrow0.5)}$  \\     \midrule \multirow{2}{*}{10\%}   &
    SVD & 84.8 & 87.8 & 86.3  & 67.0  & $94.4^{(\downarrow3.6)}$ \\
    & FMS & \textbf{85.8} & \textbf{87.9} & \textbf{86.8}  & \textbf{67.5}  & $\mathbf{97.4}^{(\downarrow0.7)}$  \\  \midrule \multirow{2}{*}{20\%}   &
    SVD & 84.5 & 87.5 & 86.0  & 66.7  & $93.6^{(\downarrow4.4)}$ \\
    & FMS & \textbf{85.6} & \textbf{88.1} & \textbf{86.8}  & \textbf{67.3}  & $\mathbf{97.2}^{(\downarrow0.9)}$  \\ 
    \bottomrule
    \end{tabular}%
\end{table}

\begin{table}[!t]
\caption{\textbf{Comparison of different low-rank subspace projection methods on a highly curved text subset of CTW1500.} Dim refers to the dimension of the basis vectors in the low-rank subspace.}
\label{tab:lrsp}
 \setlength\tabcolsep{9pt}
    \centering
    \begin{tabular}{c|c|ccccc}
    \toprule
    \multirow{2}{*}{Dim} & \multirow{2}{*}{Method} & \multicolumn{3}{c}{Detection} & \multicolumn{1}{c}{E2E} & \multirow{2}{*}{IoU}\\ 
\cmidrule(l{0.5em}r{0.5em}){3-5} 
\cmidrule(l{0.2em}r{0.2em}){6-6}
    &  & R & P & F & F & \\ 
    \midrule
      \multirow{2}{*}{6}   &  SVD & 81.1 & 84.8 & 83.0  & 61.6  & 85.3 \\
& FMS & \textbf{82.1} & \textbf{85.4} & \textbf{83.7}  & \textbf{62.4}  & \textbf{87.8} \\ \midrule
   \multirow{2}{*}{14}  & SVD &  \textbf{83.1} & 85.6 & 84.3 & 63.2 & 97.4 \\  
   & FMS & 82.9 &  \textbf{86.5} &  \textbf{84.7}  & \textbf{63.5}  & \textbf{97.9} \\
\midrule
   \multirow{2}{*}{22}  & SVD &  \textbf{83.5} & 85.1 & 84.3 & 63.3 & 99.2 \\  
   & FMS & 83.2 & \textbf{85.7} &  \textbf{84.4}  & \textbf{63.4}  & \textbf{99.4} \\

    \bottomrule
    \end{tabular}%
\end{table}

\begin{table}[!t]
\caption{\textbf{Performance of LRANet++ with different detection head designs on Total-Text.}}
\label{tab:det_head}
  \setlength\tabcolsep{3pt}
    \centering
    \begin{tabular}{c|cccccc}
    \toprule
    \multirow{2}{*}{Method} & \multicolumn{3}{c}{Detection} & \multicolumn{1}{c}{E2E}
    & \multicolumn{1}{c}{Train. Time}
    & \multicolumn{1}{c}{Infer. Time (ms) }
    \\ 
\cmidrule(l{0.5em}r{0.5em}){2-4} 
\cmidrule(l{0.2em}r{0.2em}){5-5}
    \cmidrule(l{0.2em}r{0.2em}){6-6} 
    \cmidrule(l{0.2em}r{0.2em}){7-7}
     & R & P & F & F & (s/epoch) & Det. Head \\ 
    \midrule
   Single (Dense) & 85.7 & 89.8 & 87.7  & 81.8  & \textbf{60.6}  & 29.6 \\
    Single (Sparse) & 84.6 & 89.5 & 87.0 & 81.1 & 62.0 & 6.4 \\
    Dual & \textbf{85.5} & 90.8 & \textbf{88.1} & \textbf{82.4} & 63.3 & 6.5 \\
    Triple  & 85.2 & \textbf{91.1} & \textbf{88.1} & 82.2 & 63.9 & \textbf{3.2} \\
    \bottomrule
    \end{tabular}%
\end{table}

\subsubsection{$Orthanchor$ $Space$ Dimension}

To determine the optimal $orthanchor$ $space$ dimension, denoted as $M$, we first measure the importance of each $orthanchor$ by deriving the variance of its corresponding projection coefficients:
\begin{align}\label{eq:var}
\sigma_i^2 = \mathrm{Var}(\mathbf{c}_i) = \mathrm{Var}(\mathbf{u}_i^\top \mathbf{A}).
\end{align}
A basis vector capturing higher variance corresponds to a more significant component of shape variation. As illustrated in Fig.~\ref{fig:Analysis_U}(a), the importance distribution exhibits a steep drop-off, which strongly indicates that the text shape space is inherently low-rank. This finding is mirrored in the reconstruction quality (Fig.~\ref{fig:Analysis_U}(b)), where the IoU increases sharply at lower dimensions before the gain plateaus around $M=14$.

\begin{table}[!t]
\caption{\textbf{Comparison of computational cost (FLOPs) at a unified scale (640$\times$640).} Our recognition head is omitted as its cost is dynamic since it depends on the detection output.}
\label{tab:flops}
    \centering
     \setlength{\tabcolsep}{7pt}
    \begin{tabular}{c|cccc}
    \toprule
    \multirow{2}{*}{Method} & \multicolumn{4}{c}{FLOPs (G)} \\ 
    \cmidrule(l{0.2em}r{0.2em}){2-5} 
     & Backbone  & Neck & Det. Head  & Total \\ 
    \midrule
    DB++ \cite{liao2021mask} & \textbf{24.8} & 21.3 & 8.5  & 54.6   \\
    LRANet \cite{su2024lranet} & 26.3 & \textbf{6.4} & 39.9 & 72.6 \\
    LRANet++ (det) & 26.3 & \textbf{6.4} & \textbf{1.2} & \textbf{33.9} \\
    \bottomrule
    \end{tabular}%
\end{table}

\begin{table}[!t]
 \caption{\textbf{Ablation study on the decoding strategy for our recognition head.} AR and CTC denote autoregressive and connectionist temporal classification decoding, respectively.}
 \centering
\begin{tabular}{c|c|cccc} 
    \toprule
   \multirow{2}{*}{Dataset}  & \multirow{2}{*}{Decoding} & \multicolumn{3}{c}{E2E}  & \multicolumn{1}{c}{Time (ms) }\\ 
\cmidrule(l{0.5em}r{0.5em}){3-5} 
    \cmidrule(l{0.2em}r{0.2em}){6-6} 
    & & R & P & F & Rec. Head \\  \midrule
     \multirow{2}{*}{CTW1500}
   & AR & $62.5$          & $\mathbf{75.5}$    & $\mathbf{68.4}$    & 33.2         \\
   & CTC & $\mathbf{62.9}$          & $73.3$   & $67.7$    & $\mathbf{15.1}$         \\ \midrule
\multirow{2}{*}{Total-Text} 
& AR
               & $79.1$     & $\mathbf{86.0}$    & $\mathbf{82.4}$         & $21.9$      \\  
& CTC & $\mathbf{79.2}$    & $85.4$    & $82.2$         & $\mathbf{11.1}$       \\  \bottomrule
\end{tabular}
 \label{tab:ar_vs_ctc}
\end{table}

\begin{table}[!t]
 \caption{\textbf{Effectiveness of the proposed recognition head.} Rec.: Recognition; $\dagger$ means removing the CoordConv layer, as it is time-consuming.}
 \centering
\begin{tabular}{c|c|cccc} 
    \toprule
   \multirow{2}{*}{Dataset}  & \multirow{2}{*}{Rec. Head} & \multicolumn{3}{c}{E2E}  & \multicolumn{1}{c}{Time (ms) }\\ 
\cmidrule(l{0.5em}r{0.5em}){3-5} 
    \cmidrule(l{0.2em}r{0.2em}){6-6} 
    & & R & P & F & Reg. Head\\  \midrule
     \multirow{2}{*}{CTW1500} & ABCNet v2$^{\dagger}$     & $59.8$       & $71.4$    & $65.1$    & $21.0$           \\
   & Ours & $\mathbf{62.9}$          & $73.3$    & $67.7$    & $\mathbf{15.1}$         \\ \midrule
\multirow{2}{*}{Total-Text} & ABCNet v2$^{\dagger}$   & $76.3$    & $83.9$        & $79.9$    & $17.5$      \\  
& Ours
               & $\mathbf{79.2}$     & $85.4$    & $82.2$         & $\mathbf{11.1}$       \\  \bottomrule
\end{tabular}
 \label{tab:rec.head}
\end{table}

The empirical results in Table~\ref{tab:lrsp2} are consistent with these theoretical findings. While increasing $M$ from 10 to 14, the IoU increases by 2.3\%, along with a 0.6\% increase in F-measure for both text detection and spotting. When we further increase $M$ to $18$, it yields smaller and inconsistent performance changes while increasing training complexity. Thus, we set $M=14$ to balance training complexity and representation quality.

\subsubsection{Different Low-Rank Subspace Projections}

FMS is theoretically more robust than traditional SVD to the outliers often found in real-world annotations. We conduct two experiments to verify this.

First, to validate the necessity of a robust method against the annotation errors inevitably introduced in real-world datasets, we create corrupted versions of the CTW1500 training set by injecting varying levels of spike noise (5\%, 10\%, and 20\%), where one to five random vertices are drastically displaced. We then learn the LRA basis from this noisy data. As shown in Table~\ref{tab:noise}, the $\ell_2$-based SVD is highly sensitive to outliers; even at a low noise level of 5\%, its representation quality (IoU) drops significantly by 3.0\%. As the noise ratio increases to 20\%, SVD suffers an additional 1.4\% degradation in IoU. In contrast, our FMS method exhibits better resistance to increasing noise levels. It maintains a high IoU (97.2\%) and a stable detection F-measure (86.8\%) even under severe corruption (20\% noise). This confirms that FMS extracts the intrinsic shape structure with less distortion caused by outliers, suggesting its potential for robust learning from noisy pseudo-labels in semi-supervised settings.

Second, we compare their performance on geometrically complex shapes using the highly curved text subset from \cite{zhu2021fourier}. As shown in Table~\ref{tab:lrsp}, FMS builds a higher quality projection. Crucially, at a low dimension of $M=6$, FMS achieves a significant IoU gain of 2.5\% over SVD. This indicates that the principal components derived by FMS capture the essential geometric primitives of curved text more compactly than those from SVD. This representation advantage translates to performance gains of 0.7\% and 0.8\% in detection and spotting F-measures, respectively. When the dimension increases to 14, FMS continues to outperform SVD by 0.5\% in IoU, with increases of 0.4\% and 0.3\% in F-measure for text detection and spotting, respectively. However, when the dimension increases to 22, although both FMS and SVD achieve excellent representation quality, consistent with Table~\ref{tab:lrsp2}, there is no improvement in detection performance, as the difficulty of regressing high-frequency components outweighs the marginal geometric gains.

Overall, we adopt FMS as the low-rank subspace projection method due to its superior robustness against outliers and higher representation quality with compact dimensions compared to SVD.

\begin{table}[t]
 \caption{\textbf{Effectiveness of large-ratio image scaling data augmentation on Total-Text.}}
 \centering
  \setlength\tabcolsep{12pt}
\begin{tabular}{c|ccc} 
    \toprule
    \multirow{2}{*}{Scaling Ratio} & \multicolumn{3}{c}{E2E}   \\ 
\cmidrule(l{0.5em}r{0.5em}){2-4}  
     & R & P & F  \\  \midrule
      $[0.75, 1.5]$    & 76.8    & 84.0       & 80.3            \\
    $[0.375, 2.25]$   & 78.4  & 84.9          & 81.5            \\
    $[0.1, 3]$     &  \textbf{79.2}    & \textbf{85.4}        &  \textbf{82.2}             \\   \bottomrule
\end{tabular}
 \label{tab:ratio}
\end{table}

\begin{table}[!t]
 \caption{\textbf{Performance of LRANet++ with different input sizes.}}
 \centering
  \setlength\tabcolsep{7pt}
\begin{tabular}{c|c|ccccc} 
    \toprule
    \multirow{2}{*}{Dataset} & \multirow{2}{*}{Input} & \multicolumn{3}{c}{Detection} & \multicolumn{1}{c}{E2E} & \multirow{2}{*}{FPS}\\ 
\cmidrule(l{0.5em}r{0.5em}){3-5} 
\cmidrule(l{0.2em}r{0.2em}){6-6}
    & & R & P & F & F & \\  \midrule
          & $512$    & $83.1$     & $88.1$          & $85.5$          & $64.8$   & $\mathbf{34.2}$      \\
\multirow{1}{*}{CTW1500}   & $640$  & $84.7$ & $88.7$          & $86.7$          & $66.9$     & $30.8$       \\
           & $736$   & $\mathbf{85.3}$     & $\mathbf{89.1}$      & $\mathbf{87.2}$          & $\mathbf{67.7}$    & $26.2$    \\ \midrule
\multirow{3}{*}{Total-Text}  & $608$ &  $81.5$ & $88.8$   & $85.0$                & $79.3$    & $\mathbf{34.8}$      \\
          & $800$  & $83.8$ & $90.2$    & $86.9$          & $81.2$  & $26.5$      \\ 
          & $1000$  & $\mathbf{85.2}$ & $\mathbf{91.1}$     & $\mathbf{88.1}$  & $\mathbf{82.2}$  & $20.3$     \\     \bottomrule
\end{tabular}

 \label{tab:size}
\end{table}

\begin{table}[t]
\setlength\tabcolsep{12pt}
\caption{\textbf{Comparison with different parameterized text shape methods on CTW1500.}}
\centering
\begin{tabular}{cccccc}
\hline
Method     &  Dim    &  IoU     \\ \hline
Chebyshev \cite{wang2020textray}   & $44$ & $83.6$\\ 
DCT \cite{su2022textdct}           & $32$ & $88.5$ \\ 
Fourier \cite{zhu2021fourier}           & $22$ & $91.5$\\
Bezier \cite{liu2020abcnet}       & $16$ & $97.6$\\
TPS \cite{wang2022tpsnet}           & $22$ & $97.9$\\ 
 \hline
\textbf{LRA}  & $\mathbf{14}$ & $\mathbf{98.1}$ \\ 
\hline
\end{tabular}
\label{tabel-compare-dim-iou}
\end{table}

\begin{table*}[t]
  \centering
  \footnotesize
  \caption{\textbf{Text detection results on typical benchmarks.} * denotes the results based on end-to-end text spotting training. Bold and underline refer to the first and second performances, respectively, which have the same meaning in other tables. All listed FPS values are uniformly measured using a single NVIDIA RTX 3090 GPU.}
  \resizebox{0.95\linewidth}{!}{%
    \begin{tabular}{@{}c |l |c |ccc | ccc| ccc| cccc  @{}}
      \toprule
      \multirow{2}*{Type} & \multirow{2}*{Method} & \multirow{2}*{Ext} & \multicolumn{3}{c}{MSRA-TD500} & \multicolumn{3}{c}{ICDAR 2015} & \multicolumn{3}{c}{Total-Text} &
      \multicolumn{4}{c}{CTW1500}  \\
\cmidrule(l{0.2em}r{0.2em}){4-6} 
\cmidrule(l{0.2em}r{0.2em}){7-9} 
\cmidrule(l{0.2em}r{0.2em}){10-12} 
\cmidrule(l{0.2em}r{0.2em}){13-16} 
        & & & R  & P & F & R  & P & F & R  & P & F  & R  & P & F &FPS \\
      \midrule
    \multirow{9}{*}{\rotatebox{90}{Segmentation-based}} 
   & DRRG  \cite{2020Deep}   & $\checkmark$       & $82.3$                 & $88.1$      & $85.1$    
   & $84.7$      & $88.5$   & $86.6$ 
   & $84.9$      & $86.6$   & $85.8$  & $83.0$ & $86.0$ 
    & $84.5$  & $2.0$     \\ 
     & TextBPN \cite{zhang2021adaptive} & $\checkmark$         & $84.5$                 & $86.6$      & $85.6$   & --   & --      & --  
     & $85.2$      & $90.8$       & $87.9$     & $83.6$                 & $86.5$      & $85.0$  & $18.1$         \\
     & FSG \cite{tang2022few}     & $\checkmark$     & $84.8$                 & $91.6$      & $88.1$ 
     & $87.3$                 & $90.9$      & $89.1$ 
     & $85.7$      & $90.7$       & $88.1$     & $82.4$                 & $88.1$      & $85.2$   & --         \\  
     & TextPMs \cite{zhang2022arbitrary}  & $\checkmark$        & $87.0$                 & $91.0$      & $88.9$     
      & $84.9$      & $89.9$       & $87.4$  
     & $87.7$      & $90.0$       & $88.8$    
     & $83.8$                 & $87.8$      & $85.7$    & $14.4$         \\ 
   & DB++  \cite{liao2022real}   & $\checkmark$       & $83.3$                 & $91.5$      & $87.2$ 
   & $83.9$                 & $90.9$      & $87.3$   
   & $83.2$      & $88.9$       & $86.0$     & $82.8$                 & $87.9$      & $85.3$   & $\underline{38.3}$          \\
 & TextBPN++ \cite{zhang2023arbitrary} & $\checkmark$       & $86.8$                 & \underline{93.7}      & \underline{90.1}      & --   & --      & --  
 & $87.9$      & $92.4$   & \underline{90.1}      & $84.7$                 & $88.3$      & $86.5$    & $13.9$   \\
 & CBNet \cite{zhao2024cbnet} & $\checkmark$        & $84.8$                 & $91.1$      & $87.8$   
  & --   & --      & -- 
 & $82.5$      & $90.1$       & $86.1$      & $81.9$                 & $89.0$      & $85.3$    & --     \\
   & STD  \cite{han2024spotlight}   & $\checkmark$       & $86.9$                 & $92.8$      & $89.8$
   & $85.2$      & $88.9$       & $87.0$ 
   & $83.9$      & $90.7$       & $87.2$     & $84.9$                 & $88.5$      & $86.7$   & --   \\
 & IAST \cite{zhang2024inverse}*  & --  & -- & -- & --  & 86.6 & 92.5 & 89.5 & 85.2 & \textbf{94.7}  & 89.7 & 84.8 & 89.2 & 86.9  & --
   \\ \midrule
    \multirow{14}{*}{\rotatebox{90}{Regression-based}}  
   & TextRay \cite{wang2020textray}     &  $\checkmark$     & -- & --  & --  & -- & --  & --   & $77.9$     & $83.5$      & $80.6$       & $80.4$       & $82.8$     & $81.6$      & --   \\
    & FCENet  \cite{zhu2021fourier} & --     &  --        & --  & --  & $84.2$     & $85.1$      & $84.6$    & $79.8$     & $87.4$   & $83.4$    
    & $80.7$     & $85.7$      & $83.1$  & --   \\
    & ABCNet v2 \cite{liu2021abcnet}*  & $\checkmark$             & $81.3$                 & $89.4$      & $85.2$  & $86.0$  & $90.4$      & $88.1$     & $84.1$      & $89.2$       & $87.0$  &  $83.8$                 & $85.6$      & $84.7$  & --        \\
 & TextDCT \cite{su2022textdct}   & --    & --                 & --  & --   & $83.7$     & $86.9$      & $85.3$  & $80.5$     & $85.8$      & $83.0$     & $81.5$   & $84.7$  & $83.1$    & $19.5$          \\ 
 & TPSNet \cite{wang2022tpsnet}*  & $\checkmark$          & --                 & --      & --   
 & $87.8$      & $90.5$       & $89.1$ 
 & $86.8$      & $90.2$       & $88.5$        & $86.3$        & $88.7$      & $87.5$      & $17.9$         \\ 
 & CT-Net  \cite{shao2023ct}  & --         & $80.4$        & $89.8$      & $84.8$  & $85.6$                 & $88.1$      & $86.8$  & $83.6$      & $89.2$       & $86.3$       & $82.7$     & $87.9$ 
 & $85.2$  & $13.6$           \\ 
 & DPText-DETR  \cite{ye2023dptext}   & $\checkmark$         & --                 & --      & --  & --                 & --      & --     & $86.4$      & $91.8$       & $89.0$         & $86.2$     & $91.7$ 
 & $88.8$  & $14.8$  \\
 & DeepSolo \cite{ye2023deepsolo}* & $\checkmark$ & -- & -- & -- & 87.4 & \underline{92.8} & 90.0  & 82.1 & 93.1  & 87.3 & 85.0 & \textbf{93.2} & \underline{88.9}   & 15.9 \\
      
 
 & OmniParser \cite{wan2024omniparser}*  & $\checkmark$ & -- & -- & --   & \textbf{91.0} & 90.3 & \underline{90.7}  & \underline{88.6} & 88.4  & 88.5 & \underline{87.6} & 87.9 & 87.8   & -- \\ 
  & LayoutFormer  \cite{liang2024layoutformer}   & $\checkmark$        & \underline{88.3}      & $92.0$       & \underline{90.1}  & --     & --      & --     & $85.0$      & $89.3$       & $87.1$  & $84.3$     & $88.2$ 
 & $86.2$  & --  \\  
    \cmidrule{2-16}
 
    & \textbf{LRANet} \cite{su2024lranet}       & --       & $85.3$                 & $89.1$      & $87.2$  & --   & --      & --   & $85.7$      & $90.5$       & $88.1$        & $84.9$     & $89.1$        & $86.9$   & $37.2$ \\ 
   & \textbf{LRANet++} & -- & 86.3  & 88.5 & 87.4  &  86.7  & 89.6 & 88.1 & 85.2 & 91.1  & 88.1 & 85.3 & 89.1 & 87.2 & \textbf{43.5}    \\
 & \textbf{LRANet++} & $\checkmark$  & 87.0
 & 92.8 & 89.8  &  87.3 & 91.8 & 89.5
 & 87.5 & 91.8  & 89.6  & 87.3 & 90.1  & \underline{88.9} & \textbf{43.5}    \\
 & \textbf{LRANet++*} & $\checkmark$  & \textbf{88.9}
 & \textbf{94.2} & \textbf{91.5}  & \underline{88.0}
 & \textbf{93.9} & \textbf{90.9}
 & \textbf{89.1} & \underline{92.6}  & \textbf{90.8} & \textbf{88.1} & \underline{92.8} & \textbf{90.3} & \textbf{43.5}   \\
      \bottomrule
    \end{tabular}
  }
  \label{tab:det}
\end{table*}

\begin{table*}[!t]
    \caption{ \textbf{End-to-end text spotting results on the CTW1500 dataset.}  ``\emph{S}:'' means the shorter side is fixed, and ``\emph{L}:'' means the longer side is fixed. ``None'' represents lexicon-free. ``Full'' denotes using all the words that appeared in the test set.}
    \centering
  \resizebox{0.98\linewidth}{!}{%
    \begin{tabular}{@{}l|c|l|cc|c@{}}
    \toprule
        Method & Scale & External Dataset & None & Full & FPS \\ \midrule

      TextDragon \cite{feng2019textdragon} & -- & Synth800K, Total-Text, IC15 & 39.7  & 72.4 & -- \\
      MANGO \cite{qiao2021mango} &  -- & Synth800K, Synth800K, Total-Text, IC13, IC15, COCO-Text, MLT19 & 58.9  & 78.7 & -- \\
     ABCNet v2 \cite{liu2021abcnet} &  \emph{S}: 800  & Synth150K, MLT17, Total-Text & 57.5  & 77.2 & 9.4 \\
     TESTR \cite{zhang2022text} &  \emph{S}: 1000 & Synth150K, MLT17, Total-Text & 56.0  & 81.5 & 10.3 \\
     SwinTextSpotter \cite{huang2022swintextspotter}  &  \emph{S}: 1000 & Synth150K, MLT17, Total-Text, IC13, IC15 & 51.8  & 77.0 & 2.0 \\
     TPSNet \cite{wang2022tpsnet} &  \emph{S}: 800  & Synth150K, MLT17, Total-Text & 59.7  & 79.2 & 13.8 \\
     SPTS \cite{liu2023spts} &  \emph{S}: 1000  & Synth150K, MLT17, Total-Text, IC13, IC15 & 63.6 & 83.8 & 0.5 \\
     ABINet++ \cite{fang2022abinet++} &  \emph{S}: 800 & Synth150K, MLT17, Total-Text, IC15 & 60.2  & 80.3 & 15.2 \\     
     DeepSolo \cite{ye2023deepsolo} &  \emph{L}: 1200  & Synth150K, MLT17, Total-Text, IC13, IC15 & 64.2 & 81.4 & \underline{15.9} \\
     UNITS \cite{kil2023towards} &  \emph{L}: 1920  & Synth150K, MLT17, Total-Text, IC13, IC15, TextOCR, HierText & 66.4 & 82.3 & 0.1 \\ 
     SPTS v2 \cite{liu2023spts} &  \emph{S}: 1024  & Synth150K, MLT17, Total-Text, IC13, IC15 & 63.6 & 84.3 & 4.3 \\
     ESTextSpotter \cite{huang2023estextspotter} &  \emph{S}: 800  & Synth150K, MLT19, Total-Text, IC13, IC15 & 64.9 & 83.9 & 7.3 \\
     IAST \cite{zhang2024inverse} & -- & Synth150K, MLT17, Total-Text, IC15 & 62.4 & 82.9 & -- \\
     FastTextSpotter \cite{das2025fasttextspotter} & -- & Synth150K, MLT17, Total-Text & 56.0 & 82.9 & -- \\
     OmniParser \cite{wan2024omniparser} &  --  & Synth150K, MLT17, Total-Text,  IC13, IC15, TextOCR, HierText,  COCO-Text, OI V5  & 66.8  & \underline{85.1} & -- \\
    LSGSpotter \cite{lyu2024arbitrary}  &  \emph{S}: 960  & Synth150K, MLT17, Total-Text, IC13, IC15, TextOCR  &\underline{68.9} &84.4 &7.4 \\
     \midrule
    \textbf{LRANet++} &  \emph{S}: 736  & Synth150K, MLT17, Total-Text & 67.7  & 83.8 & \textbf{26.2} \\
    \textbf{LRANet++} &  \emph{S}: 736  & Synth150K, MLT17, Total-Text, IC13, IC15, TextOCR & \textbf{70.7}  & \textbf{85.2} & \textbf{26.2} \\
    \bottomrule 
    \end{tabular} }
    \label{tab:ctw1500}
\end{table*}

\begin{table*}[!t]
    \caption{ \textbf{End-to-end text spotting results on the Inverse-Text and Total-Text datasets.}}
    \centering
  \resizebox{0.99\linewidth}{!}{%
    \begin{tabular}{@{}l|c|l|cc|cc|c@{}}
    \toprule
    \multirow{2}{*}{Method} &  \multirow{2}{*}{Scale} &  \multirow{2}{*}{External Dataset}  & \multicolumn{2}{c|}{Inverse-Text} & \multicolumn{2}{c|}{Total-Text}  & \multirow{2}{*}{FPS} \\ 
\cmidrule(l{0.2em}r{0.2em}){4-5} 
    \cmidrule(l{0.2em}r{0.2em}){6-7} 
     & & & None & Full & None & Full &  \\ 
    \midrule
      TextDragon \cite{feng2019textdragon} & -- & Synth800K, IC15 & -- & -- & 48.8  & 74.8 & -- \\   
      MANGO \cite{qiao2021mango} &  -- & Synth800K, Synth800K, IC13, IC15, COCO-Text, MLT19 & -- & -- & 72.9  & 83.6 & -- \\
     ABCNet v2 \cite{liu2021abcnet} &  \emph{S}: 1000  & Synth150K, MLT17 & 34.5 & 47.4 & 70.4  & 78.1 & 10.2 \\
     TESTR \cite{zhang2022text} &  \emph{S}: 1600 & Synth150K, MLT17  & 61.9 & 74.1 & 73.3  & 83.9 & 8.2 \\
     SwinTextSpotter \cite{huang2022swintextspotter}  &  \emph{S}: 1000 & Synth150K, MLT17, IC13, IC15 & 55.4 & 67.9 & 74.3  & 84.1 & 2.6 \\
     TTS \cite{kittenplon2022towards} & -- & Synth800K, IC13, IC15, COCO-Text, SCUT & -- & -- & 78.2 & 86.3 & -- \\
     TPSNet \cite{wang2022tpsnet} &  \emph{S}: 1000  & Synth150K, MLT17 & -- & -- & 76.1  & 82.3 & 9.8 \\
     SPTS \cite{liu2023spts} &  \emph{S}: 1000  & Synth150K, MLT17, IC13, IC15 & 38.3 & 46.2 & 74.2 & 82.4 & 0.5 \\
     ABINet++ \cite{fang2022abinet++} &  \emph{S}: 1000 & Synth150K, MLT17, IC15 & -- & -- & 77.6  & 84.5 & 11.4 \\
     DeepSolo \cite{ye2023deepsolo} & \emph{S}: 1000  & Synth150K, MLT17, IC13, IC15 & 64.6 & 71.2 & 79.7 & 87.0 & 13.2 \\
     DeepSolo \cite{ye2023deepsolo} &  \emph{S}: 1000  & Synth150K, MLT17, IC13, IC15, TextOCR & 68.8 & 75.8 & 82.5 & 88.7 & 13.2 \\
     UNITS \cite{kil2023towards} &  \emph{L}: 1920  & Synth150K, MLT17, IC13, IC15, TextOCR, HierText & -- & -- & 78.7 & 86.0 & 0.1 \\ 
     SPTS v2 \cite{liu2023spts} &  \emph{S}: 1024  & Synth150K, MLT17, IC13, IC15 & 63.4 & 74.9 & 75.5 & 84.0 & 4.4 \\
     ESTextSpotter \cite{huang2023estextspotter} &  \emph{S}: 1000  & Synth150K, MLT19, IC13, IC15 & -- & -- & 80.8 & 87.1 & 4.3 \\
     IAST \cite{zhang2024inverse} & -- & Synth150K, MLT17, IC15 & 68.8 & 80.6 & 71.9 & 83.5 & -- \\
    SwinTextSpotter v2 \cite{huang2025swintextspotter} & --  & Synth150K, MLT17, IC13, IC15  & 64.8 & 76.5 & 78.6 & 86.3 & -- \\
     OmniParser \cite{wan2024omniparser} &  --  & Synth150K, MLT17, IC13, IC15, TextOCR, HierText,  COCO-Text, OI V5  & -- & -- & \underline{84.0}  & \underline{88.9} & -- \\
     InstructOCR \cite{duan2024instructocr} &  --  & Synth150K, MLT17, IC13, IC15 & -- & -- & 77.1  & 84.1 & -- \\
    LSGSpotter \cite{lyu2024arbitrary} &  \emph{S}: 960 & Synth150K, MLT17, IC13, IC15, TextOCR  &\underline{73.7} &\underline{82.3} &81.5 &87.3 & 7.4 \\
     \midrule
    \textbf{LRANet++} &  \emph{S}: 1000  & Synth150K, MLT17 & 73.2 & 81.7 & 82.2  & 87.7 & \underline{20.3} \\
    \textbf{LRANet++} &  \emph{S}: 1000  & Synth150K, MLT17, IC13, IC15, TextOCR & \textbf{75.3}  & \textbf{83.5} & \textbf{84.6}  & \textbf{89.7} &  \textbf{20.4} \\
    \bottomrule 
    \end{tabular} }
    \label{tab:totaltext}
\end{table*}

\begin{table*}[t]
\caption{\textbf{Evaluation results on the ICDAR 2015 dataset.} ``S'', ``W'', and ``G'' denote the recognition F-measure under the ``Strong'', ``Weak'', and ``Generic'' lexicons, respectively.}
    \centering
\resizebox{.99\linewidth}{!}{
    \begin{tabular}{l|l|ccc|ccc}
    \toprule
    \multirow{2}{*}{Method} & \multirow{2}{*}{External Dataset} & \multicolumn{3}{c|}{End-to-End} & \multicolumn{3}{c}{Word Spotting} \\ 
\cmidrule(l{0.2em}r{0.2em}){3-5} 
    \cmidrule(l{0.2em}r{0.2em}){6-8} 
     & & S & W & G & S & W & G \\ 
    \midrule
     TextDragon \cite{feng2019textdragon} & Synth800K, Total-Text
    &82.5 &78.3 &65.2 &86.2 &81.6 &68.0 \\
     Mask TextSpotter V3 \cite{liao2020mask} 
    & Synth150K, MLT17, Total-Text, SCUT & 83.3 & 78,1 & 74.2 & 83.1 & 79.1 & 75.1\\
      MANGO \cite{qiao2021mango}  & Synth800K, Synth800K, Total-Text, IC13, COCO-Text, MLT19 &85.4 &80.1 &73.9 &85.2 &81.1 &74.6 \\
    ABCNet v2 \cite{liu2021abcnet} & Synth150K, MLT17, Total-Text  &82.7 &78.5 &73.0 &-- &-- &-- \\
    TESTR \cite{zhang2022text} & Synth150K, MLT17, Total-Text &85.2 &79.4 &73.6  &-- &-- &--  \\
     TTS \cite{kittenplon2022towards} & Synth800K, IC13, COCO-Text, SCUT & 85.2 & 81.7 & 77.4  & 86.3 & 82.3 & 77.3\\
    SwinTextSpotter \cite{huang2022swintextspotter} & Synth150K, MLT17, Total-Text, IC13 &83.9 &77.3 &70.5  &-- &-- &--  \\
    SRSTS \cite{wu2022decoupling} & Synth800K, Synth150K, MLT17, COCO-Text, ArT19
     &85.6 &81.7 &74.5 &85.8 &82.6 &76.8 \\
    ABINet++ \cite{fang2022abinet++} & Synth150K, MLT17, Total-Text &84.1 &80.4 &75.4 &-- &-- &-- \\
    GLASS \cite{ronen2022glass} & Synth800K &84.7 &80.1 &76.3 &86.8 &82.5 &78.8 \\
    SPTS v2 \cite{liu2023spts} & Synth150K, MLT17, Total-Text, IC13 &82.3 &77.7 &72.6 &-- &-- &-- \\
    DeepSolo \cite{ye2023deepsolo} & Synth150K, MLT17, Total-Text, IC13, TextOCR &88.0 &83.5 &79.1 &\underline{87.3} &\underline{83.8} & \underline{79.5} \\
     ESTextSpotter \cite{huang2023estextspotter}  & Synth150K, MLT19, IC13 & 87.5 & 83.0 & 78.1 &-- &-- &--  \\
    IAST \cite{zhang2024inverse} & Synth150K, MLT17, Total-Text &84.4 &80.0 &73.8 &-- &-- &-- \\ 
     FastTextSpotter \cite{das2025fasttextspotter}  & Synth150K, MLT17 & 86.6 & 81.7 & 75.4  &-- &-- &-- \\
     OmniParser \cite{wan2024omniparser} & Synth150K, MLT17, IC13, TextOCR, HierText,  COCO-Text, OI V5  & \textbf{89.6} & \textbf{84.5} & \underline{79.9} & -- & -- & -- \\
     InstructOCR \cite{duan2024instructocr} & Synth150K, MLT17, IC13 & 82.5 & 77.1 & 72.1 & -- & -- & -- 
    \\ \midrule
    \textbf{LRANet++}  & Synth150K, MLT17, Total-Text &87.0 & 82.5 &78.5 & 86.5 & 82.1  & 78.7 \\
    \textbf{LRANet++}  & Synth150K, MLT17, Total-Text, IC13, IC15, TextOCR &\underline{88.1} & \underline{84.0} &\textbf{80.2} &\textbf{87.6} & \textbf{83.9} &\textbf{80.3} \\
    \bottomrule
    \end{tabular}}
\label{tab:inverse-text}
\end{table*}

\subsubsection{Triple Assignment Detection Head}

We first motivate our multi-assignment design by analyzing the inherent limitations of two single-assignment baselines, as detailed in Table~\ref{tab:det_head}. The Single (Dense) baseline, for which we adopt the strategy from TPSNet \cite{wang2022tpsnet}, achieves a competitive detection F-measure of 87.7, but NMS is necessary to eliminate a large number of redundant predictions, which is especially time-consuming for arbitrary-shaped text instances. Conversely, the Single (Sparse) baseline resolves this speed bottleneck (6.4 ms), but its text detection and spotting performance decreases, as it lacks sufficient supervision signals to learn text shape information. Notably, the sparse assignment's training time (62.0 s/epoch) is slightly longer than that of the dense version (60.6 s/epoch), because the construction of the large-scale prediction-aware cost matrix (Eq.~\eqref{eq-Selection-metric}) is computationally intensive.

To overcome these limitations, the dual assignment scheme in LRANet \cite{su2024lranet} combined the advantages of dense assignment (sufficient supervision) and sparse assignment (fast inference). Building upon this, our proposed triple assignment strategy further refines this approach. As shown in Table~\ref{tab:det_head}, compared to the dual assignment scheme from LRANet, our triple assignment scheme reduces inference time by 3.3 ms on Total-Text. Moreover, the computational cost dramatically decreases, as demonstrated in Table~\ref{tab:flops}, with the FLOPs of the detection head dropping from 39.9G to 1.2G. This improvement is primarily due to the reduced number of convolutions (1 vs. 4) and output channels (32 vs. 256) in the head.

Notably, the inference time gain is not as pronounced as the computational scale reduction, mainly because the head inference delay shifts from a compute-bound bottleneck to a memory-bound bottleneck. This demonstrates that our simplified detection head structure successfully removed the computational bottleneck from the detection head itself. Moreover, despite the ultra-lightweight inference structure, the model's F-measure is scarcely affected owing to the self-distillation strategy, and the additional training complexity is also minimal, with an increase of only 0.6 s in training time per epoch.

\subsubsection{Different Recognition Decoding Strategies}

We compare the performance of our default CTC-based recognition head with an autoregressive (AR) variant, which incorporates an additional cross-attention layer to enable autoregressive decoding. As shown in Table~\ref{tab:ar_vs_ctc}, the AR decoder leverages its superior contextual modeling to achieve a slight accuracy advantage (a 0.7\% higher F-measure) on the long text-line dataset CTW1500. However, this performance gain becomes negligible on the word-level dataset Total-Text. This marginal gain is attributed to two key factors. First, as Fig.~\ref{fig:ESTS_vs_ABI} demonstrates, recognition accuracy is fundamentally restricted by detection inaccuracies, which cannot be overcome even by language modeling recognition heads like ABINet++ \cite{fang2022abinet++}. Second, for well-localized text regions, our Transformer-based recognition head effectively encodes contextual information, making CTC decoding sufficient for high performance. Critically, this marginal accuracy improvement comes at a significant efficiency cost. Due to its serial, character-by-character decoding nature, the inference latency of the AR decoder is substantially higher, particularly in long-text scenarios where its runtime on CTW1500 is more than double that of the CTC decoder (33.2 ms vs. 15.1 ms). Thus, we adopt CTC decoding, as it strikes a superior balance between accuracy and speed.

\subsubsection{Our Text Detection Module and Recognition Head}

We first validate the foundational ability of our text detection module by replacing the detection module in ABCNet v2 \cite{liu2021abcnet} and TPSNet \cite{wang2022tpsnet}, which have the same recognition head but differ only in the detection module. To achieve a better balance between accuracy and speed, we remove the CoordConv layer from their recognition head, as it is time-consuming. As shown in Table~\ref{tab:rec.head}, when using our detection module as the foundation, the model still achieves strong performance even when equipped with the recognition heads from ABCNet v2 and TPSNet (with the CoordConv layer removed). Compared to their original baseline performance (as shown in Tables \ref{tab:ctw1500} and \ref{tab:totaltext}), this improves the text spotting F-measure by at least 3\%. Subsequently, our recognition head outperforms the modified version of ABCNet v2 (without CoordConv) by at least 2\% in the F-measure and is 5.6 ms and 6.4 ms faster on the CTW1500 and Total-Text datasets, respectively, demonstrating its performance advantages. Notably, the recognition speed on the CTW1500 dataset is generally slower than on Total-Text, as CTW1500 is a line-level annotated dataset that requires a longer input length to achieve better performance.

\subsubsection{Large-Ratio Image Scaling Data Augmentation}

We adopt the traditional RoI operation, \emph{i.e.}, TPS alignment, to extract features for recognition. As analyzed in Sec. \ref{subsec:related-works-spotter}, RoI introduces text distortion due to the uniform sampling of feature sizes. Intuitively, this issue can be alleviated through multi-scale testing \cite{du2025svtrv2}, but multi-scale testing would prevent batch processing, which would result in a decrease in inference speed. Therefore, we adopt large-ratio image scaling data augmentation to force the model to learn and adapt to the diversity of text under various deformation conditions. As shown in Table~\ref{tab:ratio}, compared to conventional random scaling with regular ratios, applying extremely large scaling ratios (\textit{e.g.}, 0.1 to 3) to the height or width of the input image significantly improves text spotting performance, thus verifying its effectiveness in mitigating text distortion caused by fixed-size constraints.

\begin{table*}[t]
\caption{\textbf{End-to-end recognition results on RoIC13 without using lexicons.}}
\label{tab:ic13}
    \centering
\resizebox{0.9\linewidth}{!}{%
    \begin{tabular}{l|l|ccc|ccc}
    \toprule
    \multirow{2}{*}{Method} & \multirow{2}{*}{External Dataset} & \multicolumn{3}{c|}{Rotation Angle $45\degree$} & \multicolumn{3}{c}{Rotation Angle $60\degree$} \\ 
\cmidrule(l{0.2em}r{0.2em}){3-5} 
    \cmidrule(l{0.2em}r{0.2em}){6-8} 
    & & R & P & F & R & P & F \\ 
    \midrule
    CharNet \cite{xing2019convolutional} & Synth800K & 35.5 & 34.2 & 33.9 & 8.4 & 10.3 & 9.3 \\
     Mask TextSpotter V2 \cite{liao2021mask}  & Synth150K, MLT17, Total-Text, SCUT  & 45.8 & 66.4 & 54.2 & 48.3 & 68.2 & 56.6   \\
    Mask TextSpotter V3 \cite{liao2020mask} & Synth150K, MLT17, Total-Text, SCUT &68.8 & 88.5 & 76.1 & 67.6 & 88.5 & 76.6 \\
   SwinTextSpotter \cite{huang2022swintextspotter} & Synth150K, MLT17, Total-Text, IC15 & 72.5 & 83.4 & 77.6 & 72.1 & 84.6 & 77.9 
   \\
   TTS \cite{kittenplon2022towards} & Synth800K, IC13, COCO-Text, SCUT & -- & -- & 80.4 & -- & -- & 80.1 
   \\
   DeepSolo \cite{ye2023deepsolo} & Synth150K, MLT17, Total-Text, IC15 & 74.9 & 82.3 & 78.4 & 74.9 & 82.9 & 78.7 \\
   SwinTextSpotter v2 \cite{huang2025swintextspotter} & Synth150K, MLT17, Total-Text, IC15 & \underline{78.2} & 85.5 & 81.7 & \underline{79.3} & 86.3 & 82.6 
   \\ \midrule
   \textbf{LRANet++} & Synth150K, MLT17, Total-Text & 77.3 & \underline{88.6} & \underline{82.6} & 77.5 & \underline{88.8} & \underline{82.8} \\ 
   \textbf{LRANet++} & Synth150K, MLT17, Total-Text, IC15, TextOCR & \textbf{80.0} & \textbf{89.1} & \textbf{84.3} & \textbf{80.3} & \textbf{89.3} & \textbf{84.5} \\ 
    \bottomrule
    \end{tabular}}
\end{table*}

\begin{table}[!t]
 \caption{\textbf{Detection and end-to-end recognition results (1-NED) on ReCTS.}}
 \centering
  \setlength\tabcolsep{8pt}
\begin{tabular}{l|cccc} 
    \toprule \multirow{2}{*}{Method} & \multicolumn{3}{c}{Detection} & \multirow{2}{*}{1-NED}\\ 
\cmidrule(l{0.5em}r{0.5em}){2-4} 
   & R & P & F & \\  \midrule
      FOTS \cite{liu2018fots}   & $82.5$     & $78.3$ & $80.3$          & $50.8$     \\
 Mask TextSpotter V2 \cite{liao2021mask}  & $88.8$ & $89.3$  & $89.0$   & $67.8$         \\
  AE TextSpotter \cite{wang2020ae} & $91.0$ & $92.6$   & $91.8$  & $71.8$        \\
  ABCNet v2 \cite{liu2021abcnet} & $87.5$ & $93.6$   & $90.4$   & $62.7$        \\
  ABINet++ \cite{fang2022abinet++} & $89.2$ & $92.7$   & $90.9$  & $76.5$         \\
  SwinTextSpotter \cite{huang2022swintextspotter}  & $87.1$ & $94.1$   & $90.4$   & $72.5$         \\
  DeepSolo \cite{ye2023deepsolo} & $89.0$ & $92.6$   & $90.7$  & \underline{78.3}        \\
  ESTextSpotter \cite{huang2023estextspotter}
    & \textbf{91.3} & \underline{94.1}   & \textbf{92.7}   & $78.1$  \\
  SwinTextSpotter v2 \cite{huang2025swintextspotter}  & \underline{91.1} & $93.0$   & $92.1$   & $76.5$  \\ \midrule
  \textbf{LRANet++}   & $90.4$ & $\mathbf{94.2}$      & \underline{92.3}          & $\mathbf{80.3}$   \\ \bottomrule
\end{tabular}

 \label{tab:rects}
\end{table}

\subsubsection{Different Input Image Sizes}

To demonstrate the trade-off between speed and accuracy, we evaluate our model with different short side lengths. The results are shown in Table~\ref{tab:size}, revealing an increasing F-measure trend as the short side extends, accompanied by a reduction in FPS. It is worth noting that the decline in the F-measure for text detection and spotting follows similar trends, and even with FPS exceeding 34, our model still maintains high performance, achieving F-measures of 64.8\% and 79.3\% for CTW1500 and Total-Text, respectively.

\begin{figure*}[t]
\centering
\includegraphics[width=\textwidth]{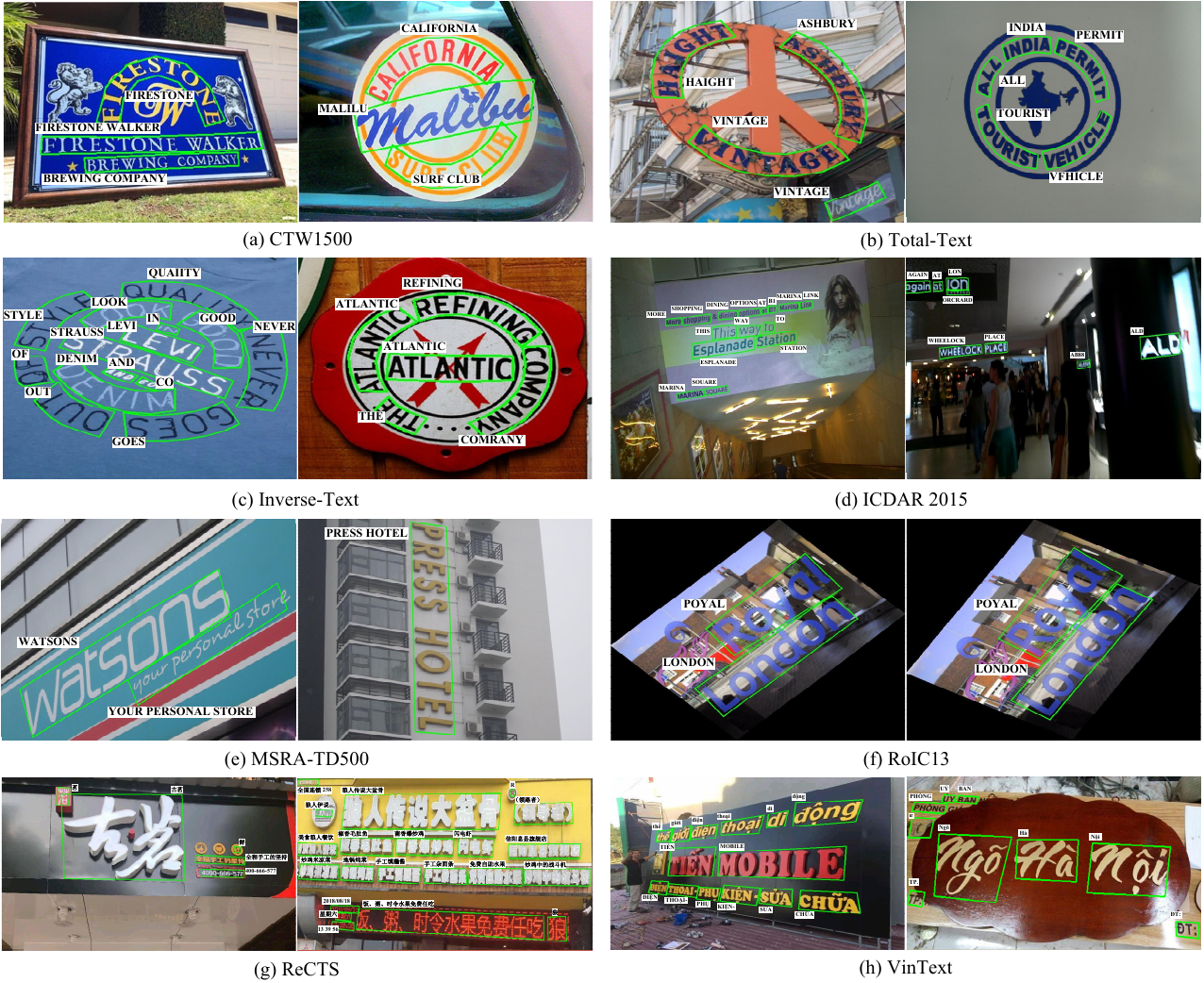}
\caption{\textbf{Visualization results of our LRANet++ on the scene text benchmarks.} Best viewed in screen.}
\label{fig:visual}
\end{figure*}

\subsection{Comparison with State-of-the-art Methods}

\subsubsection{Arbitrary-shaped Dataset}

\textbf{CTW1500.} For the line-level annotated arbitrary-shaped text benchmark CTW1500, the text detection and spotting results are presented in Table~\ref{tab:det} and Table~\ref{tab:ctw1500}, respectively. 
As illustrated in Table~\ref{tab:det}, our detection module achieves the optimal trade-off between accuracy and efficiency. Compared to DB++ \cite{liao2021mask}, the F-measure is improved by 1.9\% even without pre-training, and the speed is faster, and it is computationally efficient (33.9G vs. 54.6G, Table~\ref{tab:flops}).
When jointly trained with the recognize module, the detection performance is further improved, mainly because the gradient information from recognition helps the detector distinguish between foreground and background. Benefiting from the excellent performance of our detector, when integrated with a lightweight recognizer, it seamlessly transforms into an efficient and accurate text spotter, as shown in Table~\ref{tab:rec.head}.

As demonstrated in Table~\ref{tab:ctw1500}, our method outperforms ESTextSpotter \cite{huang2023estextspotter} by a notable margin of 2.8\% in the ``None'' metric and is 3.6 times faster. When pre-trained with TextOCR \cite{singh2021textocr}, our LRANet++ becomes the first model to exceed 70\% in the ``None'' metric, outperforms the state-of-the-art method LSGSpotter \cite{lyu2024arbitrary} by 1.9\%, and is also 3.5 times faster. These results unequivocally demonstrate its accuracy and efficiency in detecting and recognizing long-curved text. Moreover, our method outperforms OmniParser \cite{wan2024omniparser} by a significant margin of 3.9\% in the ``None'' metric, while the results are comparable on the ``Full'' metric. This discrepancy highlights a common issue in these methods: a small number of characters within a text line are often misrecognized, necessitating the use of lexicons for correction. However, in many real-world applications, lexicons are probably unavailable.

\textbf{Total-Text.} As shown in Table~\ref{tab:det}, even without pre-training on an additional scene text dataset, our method still achieves competitive detection performance. This demonstrates the strong few-shot learning capability of our approach. The text spotting results are listed in Table~\ref{tab:totaltext}, which demonstrate that our method offers advantages in inference speed, while maintaining superior performance.
As shown in Table~\ref{tab:totaltext}, at approximately the same resolution, our method is the first to exceed 20 FPS while achieving the highest accuracy. Moreover, as shown in the comparisons in Table~\ref{tab:rec.head} and Table~\ref{tab:totaltext}, when our detection module is equipped with a lightweight variant of the ABCNet v2 \cite{liu2021abcnet} recognition head, its performance significantly outperforms ABCNet v2 \cite{liu2021abcnet} and TPSNet \cite{wang2022tpsnet}, showing a 9.5\% and 3.8\% improvement in the ``None'' metric over ABCNet v2 and TPSNet, respectively. These  results highlight the effectiveness of our detection module.

\textbf{Inverse Text.} Scene text arrangements are diverse and not restricted to a left-to-right orientation. They can also appear in mirrored, symmetrical, or retroflexed layouts. To evaluate the robustness of our method in handling such arbitrary reading order text, we conduct experiments on the newly proposed Inverse-Text dataset \cite{ye2023dptext}. As shown in Table~\ref{tab:totaltext}, our method outperforms IAST \cite{zhang2024inverse} by a significant margin of 4.4\% in the ``None'' metric, although IAST is specifically designed for inverse-like scene text scenarios. Compared to LSGSpotter \cite{lyu2024arbitrary}, our method surpasses it by 1.6\% and 1.2\% in the ``None'' and ``Full'' metrics, respectively, and is 2.4x faster, despite LSGSpotter introduces a start point localization module designed to determine the reading order for inverse-like text. 

Fig.~\ref{fig:visual}(a-c) shows some qualitative results for these datasets, demonstrating the model's capability to handle long, curved, and inverted texts.

\begin{table}[!t]
 \caption{\textbf{End-to-end text spotting results on VinText.} `+D' denotes integrating the dictionary-guided method from \cite{nguyen2021dictionary}.}
 \centering
  \setlength\tabcolsep{7pt}
\begin{tabular}{l|c} 
    \toprule Method  & F-measure \\ \midrule
  ABCNet \cite{liu2020abcnet}
      & $54.2$  \\
  ABCNet+D \cite{nguyen2021dictionary}
      & $57.4$  \\
  Mask TextSpotter v3 \cite{liao2021mask}
      & $53.4$  \\
  Mask TextSpotter v3+D \cite{nguyen2021dictionary}
      & $68.5$ \\
  SwinTextSpotter \cite{huang2022swintextspotter}   & $71.1$  \\
  ESTextSpotter \cite{huang2023estextspotter}   & \underline{$73.6$} \\
  FastTextSpotter \cite{das2025fasttextspotter}   & 73.0  \\
  SwinTextSpotter v2 \cite{huang2025swintextspotter}   & $73.1$  \\ \midrule
  \textbf{LRANet++}     & $\mathbf{75.6}$    \\ \bottomrule
\end{tabular}

 \label{tab:vintext}
\end{table}

\subsubsection{Multi-Oriented Dataset}

\textbf{MSRA-TD500.} For the line-level annotated multi-oriented text detection benchmark MSRA-TD500, the results are detailed in Table~\ref{tab:det}. Our method achieves comparable performance with TextBPN++ \cite{zhang2023arbitrary} and LayoutFormer \cite{liang2024layoutformer}. When jointly trained with a recognition head, our method significantly outperforms previous methods in terms of Recall, Precision, and F-measure, achieving $88.9\%$, $94.2\%$, and $91.5\%$ for these metrics, respectively.

\textbf{ICDAR 2015.} As shown in Table~\ref{tab:det}, our method outperforms all previous methods in detection, surpassing the best-reported method, OmniParser \cite{wan2024omniparser}, by 3.6\% in Precision and 0.2\% in the F-measure, despite adopting a significantly lighter structure. In the text spotting task, our method significantly outperformed previous RoI-based methods such as ABCNet v2 \cite{liu2021abcnet} and IAST \cite{zhang2024inverse}. Furthermore, our method outperforms OmniParser in ``Generic'' metric while achieving  comparable results in ``Strong'' and ``Weak'' metrics, using significantly less training data. This highlights our method's superior ability to accurately recognize detected results and its stronger generalization capability,
as the ``Generic'' metric better reflects real-world scenarios without lexicon constraints.

\textbf{Rotated IC13 (RoIC13).} To further evaluate the rotation robustness of our method, we conduct experiments on the rotated variant of IC13 \cite{liao2021mask}. The end-to-end recognition results are shown in Table~\ref{tab:ic13}. LRANet++ achieves superior performance across all metrics on both the $45\degree$ and $60\degree$ RoIC13 datasets.

As shown in Fig.~\ref{fig:visual}(d-f), LRANet++ performs well on rotated texts. In particular, although we only train the detection module on MSRA-TD500, some recognition results in this dataset are still accurate. For example, as shown in the first column of Fig.~\ref{fig:visual}(e), even though the character ``E'' is not detected correctly, it is still recognized correctly, because the corresponding receptive field of RoI features is actually larger than the character itself.

\subsubsection{Multilingual Dataset}

\textbf{ReCTS.} Chinese text spotting is a challenge as it has thousands of character classes and complex font structures. Following \cite{huang2023estextspotter, ye2023deepsolo,fang2022abinet++,liu2021abcnet}, we evaluate Chinese text spotting performance of LRANet++ on ReCTS. As shown in Table~\ref{tab:rects}, our method sets a new state-of-the-art, achieving a 1-NED score of 80.3\% and outperforming the previous leading method, DeepSolo, by 2.0\%. This demonstrates the generalization capability of LRANet++ in handling complex, non-Latin language scenes.

\textbf{VinText.} To further test the model's multilingual capabilities, we conduct evaluations on the Vietnamese dataset VinText. As presented in Table~\ref{tab:vintext}, our method achieves a new state-of-the-art performance with an F-measure of 75.6\%, which is a 2.0\% improvement compared to the previous leading method, ESTextSpotter. It is worth noting that our method does not use the dictionary for recognition, yet it still surpasses dictionary-guided methods like MaskTextSpotter v3+D by a significant margin of 7.1\%.

Some qualitative results for these datasets are depicted in Fig.~\ref{fig:visual}(g-h). As illustrated, our method can accurately detect and recognize some of the artistic and blurry fonts shown in the figures.

\section{Analysis and Discussion}

\subsection{Generalization and Visualization Analysis of LRA}
To verify the generalization capability of LRA, we extract the first $14$ $orthanchors$ from the easily accessible synthetic dataset Synth150K \cite{liu2020abcnet} and evaluate the model performance on the long curved dataset CTW1500. As shown in Table~\ref{tab:Generalization}, compared to the $orthanchors$ extracted from their respective training datasets, $orthanchors$ obtained from Synth150K consistently maintain good representation quality and model performance on real-world scene text, demonstrating the generalization ability of our LRA.

This robust cross-dataset generalization stems from a fundamental statistical principle. Although scene text in CTW1500 (real) and Synth150K (synthetic) exhibit different superficial distributions in scale and shape, they are both relatively clean, large-scale samples of arbitrary-shaped text and thus can be viewed as being drawn from the same underlying data distribution—or a universal ``shape dictionary''—of arbitrary text contours. Based on fundamental principles of large-sample statistics, when the sample size $L$ is sufficiently large (as is the case for both datasets), the subspace $\mathbf{U}$ computed from each sample ($\mathbf{U}_{\text{CTW}}$ and $\mathbf{U}_{\text{Synth}}$) will be a highly accurate estimate that converges to the one true underlying subspace $\mathbf{U}_{\text{true}}$ of this universal dictionary. This convergence ($\mathbf{U}_{\text{CTW}} \approx \mathbf{U}_{\text{true}} \approx \mathbf{U}_{\text{Synth}}$) mathematically explains our empirical observation: since the basis vectors are almost the same, their reconstruction quality and downstream task performance on the same test set are also virtually identical.

This analysis shows that LRA accurately extracts the most significant and universal text-specific shape information. Visualization of the $orthanchors$ in Fig.~\ref{fig:architecture} further supports this observation. As illustrated, the first few $orthanchors$, which capture the most variance (see Fig.~\ref{fig:Analysis_U}(a)), primarily model the overall information of the text shape, such as the rough outline, curvature state, and orientation. The subsequent $orthanchors$ progressively focus on capturing more localized details and complex shape variations. This hierarchical ``coarse-to-fine'' structure visually confirms the existence of a common shape dictionary, where the most critical basis vectors represent universal shape primitives shared across all text datasets.

\subsection{Running Time Analysis} 
Table~\ref{tab:time_cost} presents the time cost distribution  of all components in LRANet++. We observe that the text recognition component accounts for at least 40\% of the total time cost. Owing to the modular design of our LRANet++, we optimize the speed by decoupling the recognition module through a standard producer-consumer pipeline. This parallelization approach can reduce the time cost of recognition to half of its original value, leading to a significant overall speed improvement, as shown in Row 1 of Table~\ref{tab:time_cost}. Moreover, as shown in Table~\ref{tab:flops}, the neck and detection head modules introduce minimal computational latency, which is crucial for achieving real-time text spotting in our method.

\begin{table}[!t]
\caption{\textbf{Generalization Evaluation of LRA on CTW1500.}}
\label{tab:Generalization}
    \centering
    \begin{tabular}{c|ccccc}
    \toprule
    \multirow{2}{*}{Orthanchors Space } & \multicolumn{3}{c}{Detection } & \multicolumn{1}{c}{E2E} & \multirow{2}{*}{IoU}\\ 
\cmidrule(l{0.5em}r{0.5em}){2-4} 
\cmidrule(l{0.2em}r{0.2em}){5-5}
     & R & P & F & F & \\ 
    \midrule
    CTW1500 & \textbf{85.3} & 89.1 & \textbf{87.2} & \textbf{67.7}  & \textbf{98.1} \\
    Synth150K  & 84.9 & \textbf{89.4} & 87.1 & \textbf{67.7} & $98.0^{(\downarrow0.1)}$  \\
    \bottomrule
    \end{tabular}%
\end{table}

\begin{table}[!t]
\caption{\textbf{Time cost of LRANet++ on the CTW1500 dataset.} 
Det.: Detection. Rec.: Recognition. $\ddagger$ means that we adopt a producer-consumer pipeline to perform recognition in parallel.}
\label{tab:time_cost}
    \centering
     \setlength{\tabcolsep}{4pt}
    \begin{tabular}{c|cccc|c}
    \toprule
    \multirow{2}{*}{Method} & \multicolumn{4}{c|}{Time consumption (ms)}  & \multirow{2}{*}{FPS}\\ 
    \cmidrule(l{0.2em}r{0.2em}){2-5} 
     & Backbone  & Neck & Det. Head & Rec. Head & \\ 
    \midrule
    LRANet++ 512$^{\ddagger}$ & 10.6 & 1.9 & 2.5  & 7.6  & 44.2 \\
    LRANet++ 512 & 10.6 & 1.9 & 2.5 & 14.2  & 34.2 \\
    LRANet++ 640 & 11.8 & 3.1 & 2.8 & 14.9 & 30.8 \\
    LRANet++ 736  & 16.0 & 4.3 & 2.9 & 15.1 & 26.2 \\
    \bottomrule
    \end{tabular}%
\end{table}

\subsection{Performance Analysis}

We further analyze why LRANet++ achieves such accurate and efficient results. We choose TPSNet \cite{wang2022tpsnet} as our baseline, as it shares the same backbone, neck, and RoI-based recognition paradigm with our method, allowing for a focused comparison.

In the detection module, LRANet++ and TPSNet differ only in the detection head and the text shape representation. Regarding the head, compared to the dense assignment of TPSNet, our design is not only substantially faster (3.2 ms vs. 29.6 ms) by alleviating the NMS bottleneck, but also more accurate (88.1\% vs. 87.7\% F-measure) due to the better supervision from our dynamic positive sampling strategy (Table~\ref{tab:det_head}). In terms of shape representation, our LRA achieves higher representation quality (Table~\ref{tabel-compare-dim-iou}) with significantly fewer parameters (14 vs 22). As seen in Table~\ref{tab:det}, our detector consistently outperforms TPSNet across the benchmarks, even though TPSNet employs an additional border alignment loss.

Building on this strong detection foundation, our well-designed spotting pipeline further enhances the end-to-end performance. As quantified in Table~\ref{tab:ratio}, our large-scale scaling data augmentation strategy effectively mitigates the feature distortion issue caused by RoI operations, leading to a significant 1.9\% improvement in the final spotting F-measure. Additionally, our Transformer-based recognition head is superior to the one used by TPSNet (\emph{i.e.}, the recognition head from ABCNet v2 \cite{liu2021abcnet}). As shown in Table~\ref{tab:rec.head}, our recognizer delivers a 2.6\% F-measure gain on CTW1500, rising from 65.1\% to 67.7\%, while also reducing inference time from 21.0 ms to 15.1 ms.

In summary, the superior performance of LRANet++ stems from its efficient and accurate detection foundation, and the well-designed overall architecture built upon it. This also validates the core design philosophy we proposed in our introduction.

\begin{figure}[t]
    \centering
    \includegraphics[width=0.98\linewidth]{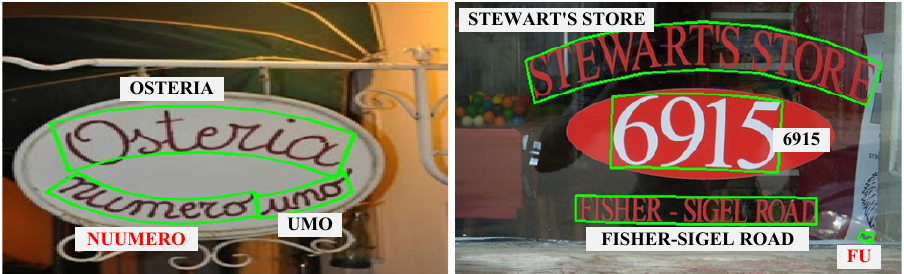}
    \begin{minipage}[t]{0.48\linewidth}
        \vspace{-0.3cm}
        \centering
        \footnotesize
       (a) Cursive Text
    \end{minipage}
    \hfill 
    \begin{minipage}[t]{0.48\linewidth}
        \vspace{-0.3cm}
        \centering
        \footnotesize
        (b) Blurry Small Text
    \end{minipage}

    \includegraphics[width=0.98\linewidth]{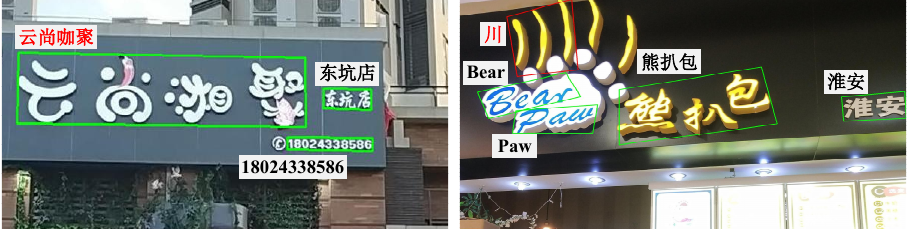}
    \begin{minipage}[t]{0.48\linewidth}
        \vspace{-0.3cm}
        \centering
        \footnotesize
        (c) Artistic Font
    \end{minipage}
    \hfill
    \begin{minipage}[t]{0.48\linewidth}
        \vspace{-0.3cm}
        \centering
        \footnotesize
        (d) Text-like Background
    \end{minipage}

    \includegraphics[width=0.98\linewidth]{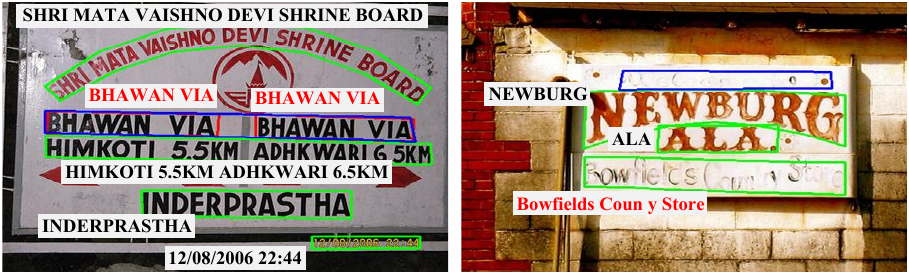}
    \begin{minipage}[t]{0.48\linewidth}
        \vspace{-0.3cm}
        \centering
        \footnotesize
        (e) Fragmented Text Line
    \end{minipage}
    \hfill
    \begin{minipage}[t]{0.48\linewidth}
        \vspace{-0.3cm}
        \centering
        \footnotesize
        (f) Overexposed Text
    \end{minipage}
    \caption{\textbf{Visualization of failure cases of our method.} Correctly predicted results are shown in black and incorrect ones in red; blue contours represent the GT.}
    \label{fig:weakness}
\end{figure}

\subsection{Failure Cases and Discussion}

As demonstrated in the experimental results, the proposed LRANet++ performs well in most cases of arbitrary-shaped text spotting. However, failure cases may still arise due to the high complexity of scene images. For example, the model may produce incorrect results when dealing with highly stylized fonts, such as the complex cursive in Fig.~\ref{fig:weakness}(a) and the intricate artistic font in Fig.~\ref{fig:weakness}(c). Similarly, blurry small text instances often lack the necessary visual detail for accurate recognition (Fig.~\ref{fig:weakness}(b)), a problem often exacerbated by such text being labeled as ``Don’t Care'' during training. Moreover, the model can occasionally be confused by text-like background textures, resulting in false positives (Fig.~\ref{fig:weakness}(d)), or it may incorrectly fragment a single line of text into multiple instances (Fig.~\ref{fig:weakness}(e)). Finally, extreme lighting conditions like overexposure can wash out pixel information, leading to missed detections or misrecognition (Fig.~\ref{fig:weakness}(f)).

These cases are common challenges for text spotting systems that require further research. At the data level, expanding the scale and diversity of training data remains a key approach to improve robustness. Architecturally, developing unified frameworks for multi-granularity spotting could enhance the model's understanding of text cohesion, as well as developing specialized backbones for more robust scene text feature extraction.

\section{Conclusion}
\label{sec:conclude}

In this paper, we have presented LRANet++, a real-time end-to-end text spotter. We first build an accurate and efficient detection foundation, featuring a low-rank subspace vector-based reconstruction to effectively leverage text-specific shape information. It accurately regresses the text contour with fewer parameters. Meanwhile, we propose a triple assignment detection head that integrates dense and sparse assignments with a self-distillation strategy for efficient inference. Based on this detection foundation, we thoroughly analyze its design for real-time text spotting, achieving a renaissance of RoI-based text spotting methods. In particular, we propose a progressive Transformer-based lightweight recognition module for efficient and accurate text recognition, along with large-scale image data augmentation to accommodate text diversity. Experiments conducted on public benchmarks basically verify the proposed LRANet++, where top-ranked accuracy and real-time inference speed are simultaneously observed. We hope that our LRANet++ can serve as a fundamental tool for many real-world text understanding tasks.

{
\bibliographystyle{ieeetr}
 \bibliography{CSRef}
}

\vfill

\end{document}